%% file: egpaper.tex
\documentclass[10pt,twocolumn,letterpaper]{article}

\usepackage{wacv}
\usepackage{times}
\usepackage{epsfig}
\usepackage{graphicx}
\usepackage{amsmath}
\usepackage{amssymb}
\usepackage{booktabs}
\usepackage{url}
\usepackage{amsfonts}       
\usepackage{nicefrac}       
\usepackage{microtype}
\usepackage{xcolor}         
\usepackage{times}
\usepackage{epsfig}
\usepackage{pifont}
\usepackage{kotex}
\usepackage{enumitem}
\usepackage{verbatim}
\usepackage{amsthm}
\usepackage{bm}
\usepackage{makecell,multirow, tabularx}
\usepackage{algorithm}
\usepackage{algorithmic}
\usepackage{url}
\usepackage{balance}
\usepackage{upquote}
\usepackage{appendix}
\usepackage{nicefrac}
\usepackage{physics}
\usepackage{stmaryrd}
\usepackage{pifont}
\usepackage{wrapfig} 
\DeclareMathOperator*{\argmax}{arg\,max}

\usepackage[accsupp]{axessibility}  

\usepackage[caption=false]{subfig}
\def\wacvPaperID{327} 

\wacvalgorithmstrack   

\wacvfinalcopy 

\ifwacvfinal
\usepackage[breaklinks=true,bookmarks=false]{hyperref}
\else
\usepackage[pagebackref=true,breaklinks=true,colorlinks,bookmarks=false]{hyperref}
\fi

\begin{document}

\title{Inducing Data Amplification Using Auxiliary Datasets in Adversarial Training}

\author{Saehyung Lee\thanks{Correspondence to: Saehyung Lee \href{mailto:halo8218@snu.ac.kr}{halo8218@snu.ac.kr}.}~~~~~~~ Hyungyu Lee\\
Department of Electric and Computer Engineering\\
Seoul National University, Seoul 08826, South Korea\\
{\tt\small \{halo8218, rucy74\}@snu.ac.kr}
}

\maketitle
\thispagestyle{empty}

\input{abstract}
\input{introduction}
\input{Methods}
\input{exp}
\input{conclusion}
\paragraph{Acknowledgements:}
This work was supported by the BK21 FOUR program of the Education and Research Program for Future ICT Pioneers, Seoul National University in 2022.
\balance

{\small
\bibliographystyle{ieee_fullname}
\bibliography{egbib}
}

\newpage
\nobalance
\appendix
\input{appendix}
\end{document}

%% file: abstract.tex
\begin{abstract}
Several recent studies have shown that the use of extra in-distribution data can lead to a high level of adversarial robustness. However, there is no guarantee that it will always be possible to obtain sufficient extra data for a selected dataset. In this paper, we propose a biased multi-domain adversarial training (BiaMAT) method that induces training data amplification on a primary dataset using publicly available auxiliary datasets, without requiring the class distribution match between the primary and auxiliary datasets. The proposed method can achieve increased adversarial robustness on a primary dataset by leveraging auxiliary datasets via multi-domain learning. Specifically, data amplification on both robust and non-robust features can be accomplished through the application of BiaMAT as demonstrated through a theoretical and empirical analysis.
Moreover, we demonstrate that while existing methods are vulnerable to negative transfer due to the distributional discrepancy between auxiliary and primary data, the proposed method enables neural networks to flexibly leverage diverse image datasets for adversarial training by successfully handling the domain discrepancy through the application of a confidence-based selection strategy. The pre-trained models and code are available at: \url{https://github.com/Saehyung-Lee/BiaMAT}.
   
\end{abstract}

%% file: introduction.tex
\section{Introduction}

The usefulness of adversarial examples in training deep neural networks (DNNs) demonstrates that the method through which these structures perceive the world is markedly different from that employed by humans.
Many approaches~\cite{adv_survey} have been proposed to bridge the gap in adversarial robustness between humans and DNNs.
Among these, training based on the use of adversarial examples as training data is considered as the most effective method to improve the robustness of DNNs.
Unfortunately, as demonstrated by Schmidt~et~al.~\cite{more_data_adversarial}, the sample complexity of adversarially robust generalization is substantially higher than that of standard generalization. To address this issue, several recent studies~\cite{unlabeled_adv,are_labels_adv} leveraged extra (in-distribution) unlabeled data and developed methods for improving the sample complexity of robust generalization. However, although such methods enable state-of-the-art adversarial robustness, they are not always capable of obtaining extra in-distribution data for any selected data distribution.
In this paper, we propose a biased multi-domain adversarial training (BiaMAT) method to improve the adversarial robustness of a classifier on a primary dataset based on the use of publicly available (labeled) auxiliary datasets. The proposed method yields the desired effect based on the following assumption:
\newtheorem{assumption}{Assumption}
\begin{assumption} \label{assumption}
    A common robust and non-robust feature space exists between the primary and auxiliary datasets.
\end{assumption}
\input{Figure/fig_rob_nonrob}
Figure~\ref{fig_rob_nonrob} shows that robust features~\cite{not_bugs_are_features} exhibit human-perceptible patterns. We may assume that if two datasets are similar from a human perspective, then they share robust features. However, non-robust features are imperceptible to humans, thus, we cannot determine whether Assumption~\ref{assumption} is correct by a human perception.
Fortunately, recent studies~\cite{transferability_adv,oat} have provided empirical evidence in support of the presence of a common non-robust feature space among diverse image datasets.
Therefore, unlike existing state-of-the-art methods~\cite{unlabeled_adv,are_labels_adv}, which employ in-distribution data, under BiaMAT, the distribution of the auxiliary dataset and the corresponding primary dataset can differ. For example, by applying BiaMAT, we can leverage CIFAR-100~\cite{cifar_dataset}, Places365~\cite{places365}, or ImageNet~\cite{downsampled_imagenet,imagenet_dataset} as an auxiliary dataset for adversarial training on CIFAR-10~\cite{cifar_dataset}.

The proposed method achieves an inductive transfer between adversarial training on the primary dataset~(referred to as the ``primary task'') and auxiliary dataset~(referred to as ``auxiliary task''). In other words, BiaMAT learns primary and auxiliary tasks in parallel within the framework of multi-domain learning~\cite{multidomain1}, and the inductive bias provided by the auxiliary tasks is transferred to the primary task through a common hidden structure. 
This mechanism can be considered to be an increase in the size of the training dataset~\cite{caruana1997multitask}. In addition,
based on studies that have demonstrated the presence of non-robust features \cite{odds_with_accuracy,not_bugs_are_features}, we classify the effects of adversarial training into two types and demonstrate the usefulness of the proposed method irrespective of the type considered. 
In particular, we dissociate the compound effect of the proposed method into the effects of \emph{non-robust feature regularization} and \emph{robust feature learning} and assess the contribution of each through the use of the expectation of random labels~\cite{caruana1997multitask,oat}. 
Our experimental results on the CIFAR datasets and ImageNet demonstrate that BiaMAT can effectively use training signals generated from various auxiliary datasets.
Furthermore, we show that while existing methods are vulnerable to negative transfer due to the distributional discrepancy between auxiliary and primary data, the proposed method enables neural networks to flexibly leverage diverse image datasets for adversarial training by successfully dealing with domain discrepancy through the application of a confidence-based selection strategy.

%% file: Figure/fig_rob_nonrob.tex
\begin{figure}[t]
\begin{center}
\includegraphics[width=\linewidth]{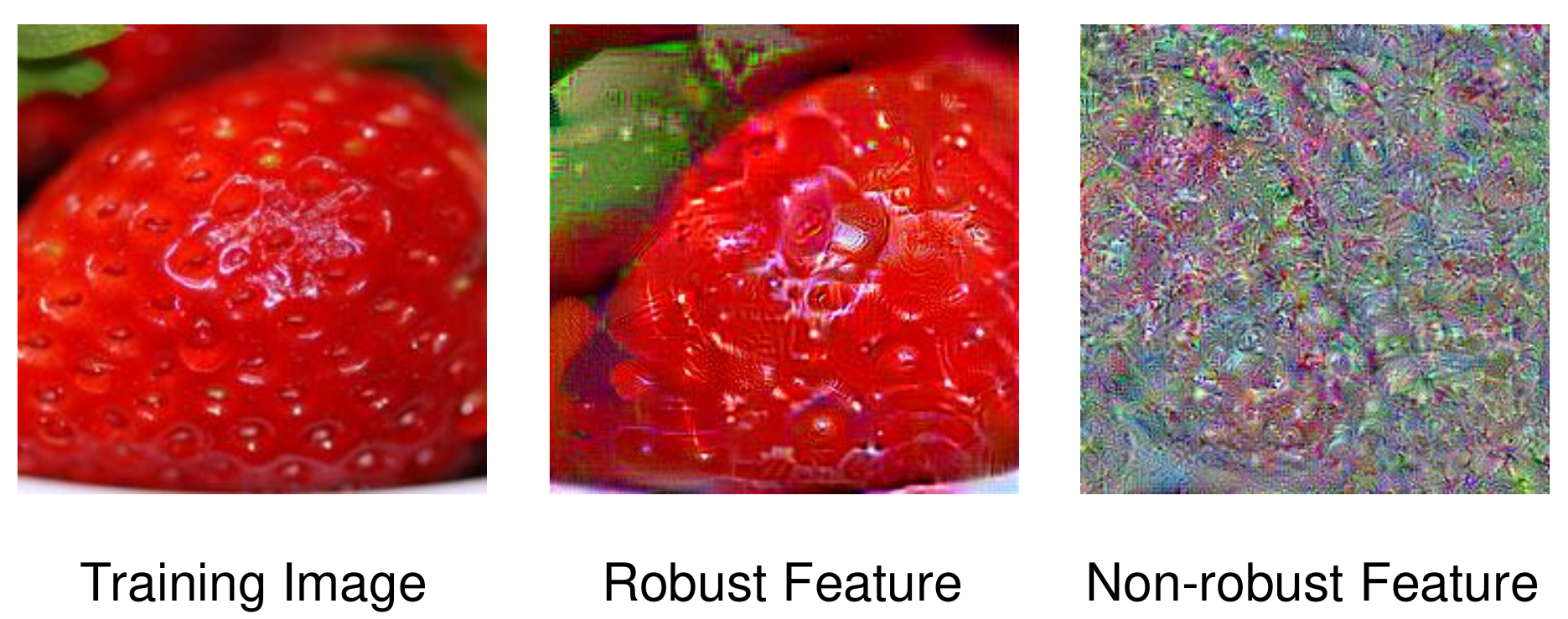}
\end{center}
   \caption{Visualization of robust and non-robust features~\cite{visrobnonrob,not_bugs_are_features}}
\label{fig_rob_nonrob}
\end{figure}

%% file: Methods.tex
\section{Biased multi-domain adversarial training}\label{sec:methods}
\subsection{Method}\label{subsec:algorithm}
\input{Figure/fig_main} 
Existing state-of-the-art methods leveraging extra unlabeled data \cite{unlabeled_adv,are_labels_adv} first remove out-of-distribution (OOD) data from a given auxiliary dataset, and then use the remaining data for training with pseudo-labels. Therefore, they are only effective when a given auxiliary dataset contains a large number of in-distribution data, and are suboptimal in terms of data utility. To maximize the data utility, we leverage given auxiliary data via multi-domain learning~\cite{multidomain2}.
Multi-domain learning is a strategy for improving the performance of tasks that solve the same problem across multiple domains by sharing information across these domains.
In standard settings, domains typically share semantic features. For example, in the Office dataset~\cite{office}, data that belong to the same class (\textit{e.g.}, keyboard) are separated into different domains (\textit{e.g.}, amazon, webcam). In adversarial settings, by contrast, it is possible to find evidence for tighter-than-expected relationships between different datasets~\cite{transferability_adv}. In particular, the use of domain-agnostic adversarial examples~\cite{transferability_adv} and robust training methods that leverage different datasets~\cite{Alvin,oat} demonstrates that common adversarial spaces can exist across different datasets. Therefore, our proposed method expands the range of related domains relative to that considered under standard settings with the goal of \emph{maximizing the adversarial robustness of the classifier on one primary dataset}. In this respect, BiaMAT differs from standard multi-domain learning, for which the primary goal is increasing the average performance over multiple domains.

We classify given auxiliary data into two types: (\romannumeral 1)~auxiliary data that share robust and non-robust features \cite{not_bugs_are_features} with the primary dataset; and (\romannumeral 2)~that share only non-robust features with the primary dataset.
We do not consider auxiliary data sharing only robust features with the primary dataset based on studies demonstrating the presence of a common non-robust feature space among
diverse image datasets \cite{adv_transfer,oat}.
As previous multi-domain learning studies observed~\cite{caruana1997multitask,multidomain1}, using (\romannumeral 1) will be beneficial to the primary task performances.
By contrast, (\romannumeral 2) can help the primary task by regularizing the shared non-robust features but, at the same time, it can suppress the advantages of multi-domain learning by generating an inductive bias toward extraneous robust features--an effect, called ``negative transfer''. Therefore, we need a method that can acquire training signals for non-robust features without achieving inductive bias for robust features from (\romannumeral 2). In Sec.~\ref{Methods: theory}, we theoretically demonstrate that for each of the two data types, multi-domain learning can improve the adversarial robustness of the primary task in a simple Gaussian model. Especially, in the theoretical case of infinite batch size, we demonstrate that the adversarial training on (\romannumeral 2) with random labels can improve adversarial robustness while avoiding negative transfer. Based on our theoretical analysis, we propose the use of the expectation of random labels ($y^{\textup{ER}}$) on (\romannumeral 2). Interestingly, the use of $y^{\textup{ER}}$ is identical to the recently proposed OAT method \cite{oat}, which uses OOD data to improve adversarial robustness.
We compare our method with OAT in Appendix~\ref{append:comp}.
To classify given auxiliary data into the two types, our proposed method utilizes a confidence-based data selection strategy. Confidence scores are used in many research fields, including semi-supervised learning \cite{fixmatch,rizve2021in} and OOD detection \cite{outlier}. For example, in the case of semi-supervised learning, high-confidence unlabeled data are used with their pseudo-labels, while the remaining unlabeled data are not used. By contrast, in our proposed method, the use of $y^{\textup{ER}}$ naturally separates the auxiliary data according to their robust features, and adversarial training algorithms for each of the two data types are applied. We provide the detailed description of the confidence-based selection strategy in Sec.~\ref{Methods: confidence-based approach}.

To sum up, we consider a multi-domain learning problem on a primary dataset $D\subset\mathcal{X}\times\mathcal{Y}$ and an auxiliary dataset $\tilde{D}\subset\mathcal{X}\times\tilde{\mathcal{Y}}$. $\tilde{D}$ is divided into two data subsets $\tilde{D}_{\textup{high}}$ (high confidence) and $\tilde{D}_{\textup{low}}$ (low confidence) through the confidence-based selection strategy. The hypotheses of the tasks are denoted by $h_\textup{pri}:\mathcal{X}\to\mathcal{Y}$ and $h_\textup{aux}:\mathcal{X}\to\tilde{\mathcal{Y}}$, where $h_\textup{pri}=f_\textup{pri}\circ g$ and $h_\textup{aux}=f_\textup{aux}\circ g$. Here, $f_\textup{pri}$ and $f_\textup{aux}$ are the prediction functions that output class probabilities for the primary and auxiliary datasets, respectively, and $g$ is the shared feature embedding function.
The loss function for $D$ is defined as $\ell=\mathop{\mathbb{E}}_{(\bm{\mathnormal{x}},\mathnormal{y})\mathtt{\sim}D} \left[\ell_{\textup{adv}}(\bm{\mathnormal{x}},\mathnormal{y};h_\textup{pri},S)\right]$, where $\ell_{\textup{adv}}$ is the adversarial loss, and $S$ represents the set of perturbations an adversary can apply. Any existing adversarial losses~\cite{pgd_attack,trades_defense} can be employed for $\ell_\textup{adv}$.
In addition, the loss for $\tilde{D}$ is defined as $\tilde{\ell}=\frac{1}{|\tilde{D}|}\sum_{(\tilde{\bm{\mathnormal{x}}},\tilde{\mathnormal{y}})\mathtt{\sim}\tilde{D}_{\textup{high}}} \ell_{\textup{adv}}(\tilde{\bm{\mathnormal{x}}},\tilde{\mathnormal{y}};h_\textup{aux},S)
+\frac{1}{|\tilde{D}|}\sum_{(\tilde{\bm{\mathnormal{x}}},\tilde{\mathnormal{y}})\mathtt{\sim}\tilde{D}_{\textup{low}}} \ell_{\textup{adv}}(\tilde{\bm{\mathnormal{x}}},\mathnormal{y}^\textup{ER};h_\textup{pri},S)$.
Our goal is to attain a small adversarial loss on the primary dataset. Thus, the proposed method minimizes the following loss:
\begin{equation} \label{eq:biamat}
    \mathcal{L}=\ell+\alpha\tilde{\ell},
    \quad\textup{where}\;\alpha\in[0,1].
\end{equation}
$\alpha$ is a hyperparameter that biases the multi-domain learning toward the primary task. Although we describe BiaMAT at the dataset level, we actually apply BiaMAT at the mini-batch level. Figure~\ref{fig_main} provides an overview of BiaMAT.

\subsection{Theoretical motivation}\label{Methods: theory}
We analyze the proposed method from the perspective of \emph{non-robust feature regularization} and \emph{robust feature learning}, which are the two effects of adversarial training.
In particular,
we define a simple Gaussian model to demonstrate how the proposed method induces training data amplification using an auxiliary dataset that satisfies Assumption~\ref{assumption}.
\paragraph{Preliminary.}\label{background:roust features}
Tsipras et al.~\cite{odds_with_accuracy} described the effect of adversarial training by constructing a classification task through which training examples $(\bm{\mathnormal{x}}, \mathnormal{y})\in\mathbb{R}^{d+1}\times\{\pm1\}$ are drawn from a distribution, as follows:
\begin{equation} \label{eq:prev data model}
\begin{gathered}
    \mathnormal{y}\stackrel{u.a.r}{\sim} \{-1, +1\},
    \quad\mathnormal{x}_1=
    \begin{cases}
    +\mathnormal{y}&\textup{w.p.}\;\mathnormal{p}\\
    -\mathnormal{y}&\textup{w.p.}\; 1-\mathnormal{p}
    \end{cases},\\
    \quad\mathnormal{x}_2,\dots,\mathnormal{x}_{d+1}\stackrel{i.i.d.}{\sim}\mathcal{N}(\eta\mathnormal{y}, 1),
\end{gathered}
\end{equation}
where $\mathnormal{x}_1$ is a robust feature that robustly correlates to the label~($\mathnormal{p}\geq0.5$), and the remaining features $\mathnormal{x}_2,\dots,\mathnormal{x}_{d+1}$ are non-robust features that are vulnerable to adversarial attacks~($0<\eta<\ell_\infty\text-$bound).
For this data distribution, the authors demonstrated that the following linear classifier could attain a standard accuracy arbitrarily close to 100\%, although it is susceptible to adversarial attacks:
\begin{equation} \label{eq:linear classifier}
    f(\bm{\mathnormal{x}})=\textup{sign}(\bm{\mathnormal{w}_{\textup{unif}}}^\top\bm{\mathnormal{x}}),
    \: \textup{where}\;\bm{\mathnormal{w}}_{\textup{unif}}=\left[ 0,\frac{1}{d},\dots,\frac{1}{d} \right].
\end{equation}
Notably, the following lemma indicates the importance of adversarial training:
\newtheorem{lemma}{Lemma}
\begin{lemma}\label{lemma1}
\textup{(Tsipras~et~al.)} Adversarial training results in a classifier that assigns zero weight to non-robust features~$\mathnormal{x}_2,\dots,\mathnormal{x}_{d+1}$.
\end{lemma}
Lemma~\ref{lemma1} shows that adversarial training (\romannumeral 1)~lowers the sensitivity of the classifier to non-robust features and (\romannumeral 2)~achieves a certain level of classification accuracy by learning robust features. We refer to (\romannumeral 1) and (\romannumeral 2) as \emph{non-robust feature regularization} and \emph{robust feature learning}, respectively.
\paragraph{Setup and overview.}
Given a shared feature embedding function $g:\mathcal{X} \to \mathcal{Z}$, we define primary and auxiliary data models in the feature space $\mathcal{Z}$, sampled from each of the following distributions:
\begin{equation}\label{datamodel}
\begin{gathered}
    \textup{(Primary)}\quad\mathnormal{y}\stackrel{u.a.r}{\sim} \{-1, +1\},\quad
    \mathnormal{z}_1\stackrel{}{\sim}\mathcal{N}(y,\mathnormal{u}^2),
    \\\mathnormal{z}_2,\dots,\mathnormal{z}_{d+1}\stackrel{i.i.d.}{\sim}\mathcal{N}(\eta\mathnormal{y}, 1),\\
    \textup{(Auxiliary)}\quad\tilde{\mathnormal{y}}=\textup{sign}(\gamma)\cdot\mathnormal{y},\quad
    \tilde{\mathnormal{z}}_1\stackrel{}{\sim}\mathcal{N}(y\abs{\gamma},\mathnormal{v}^2),
    \\\tilde{\mathnormal{z}}_2,\dots,\tilde{\mathnormal{z}}_{d+1}\stackrel{i.i.d.}{\sim}\mathcal{N}(\eta\mathnormal{y}\abs{\gamma}, 1),
\end{gathered}
\end{equation}
where $\gamma\in[-1,1]$ is the correlation coefficient between the two tasks.
We use the accent~(tilde) to represent variables associated with the ``auxiliary task''.
In Eq.~(\ref{datamodel}), $\mathnormal{z}_1$ is a robust feature that robustly correlates to the label, whereas the other features $\mathnormal{z}_2,\cdots,\mathnormal{z}_{d+1}$ are non-robust features that are vulnerable to adversarial attacks~($0<\eta<\ell_\infty\text-$bound).
If the two datasets are highly correlated in terms of robust and non-robust features, from Eq.~(\ref{eq:linear classifier}), it is evident that the following linear classifiers can achieve a high standard accuracy on the primary and auxiliary datasets, respectively, although they have low adversarial robustness:
\begin{equation}
    \begin{gathered} \label{our_classifier}
        \textup{(Primary)}\quad p(\mathnormal{y}\mid\bm{\mathnormal{z}})=\frac{1-\mathnormal{y}}{2}+\mathnormal{y}\sigma(\bm{\mathnormal{w}}^\top\bm{\mathnormal{z}}),\\
        \textup{(Auxiliary)}\quad p(\tilde{\mathnormal{y}}\mid\tilde{\bm{\mathnormal{z}}})=\frac{1-\tilde{\mathnormal{y}}}{2}+\tilde{\mathnormal{y}}\sigma(\gamma\bm{\mathnormal{w}}^\top\tilde{\bm{\mathnormal{z}}}),
    \end{gathered}
\end{equation}
where $\bm{\mathnormal{w}}=\left[ 0,\frac{1}{d},\dots,\frac{1}{d} \right]$, and $\sigma(\cdot)$ denotes a sigmoid function.
To study the effect of adversarial training on the auxiliary task on the non-robust feature regularization for the primary task, we derive the gradients of the primary and auxiliary adversarial losses with respect to the non-robust features, which are then back-propagated through the shared feature embedding function $g$. In addition, we demonstrate how the use of random labels enables us to dissociate the compound effect of the proposed method into the effects of non-robust feature regularization and robust feature learning.
\paragraph{Non-robust feature regularization.}
First, we generate the adversarial feature vector of our classification model ($\tilde{\bm{\mathnormal{z}}}^{\textup{adv}}=g(\tilde{\bm{\mathnormal{x}}}+\bm\delta):\bm\delta\in S$, where $\tilde{\bm{\mathnormal{x}}}\in\mathcal{X}$ denotes the auxiliary input vector).
The objective function of the adversary to deceive our model is the cross-entropy loss~\cite{pgd_attack}.
\begin{lemma}\label{lemma2}
Let $i\in\{2,\cdots,d+1\}$ and $\eta<\lambda<1$. Then, the expectation of the adversarial feature vector against the auxiliary task is
\begin{equation}\label{eq:lemma2}
\mathop{\mathbb{E}}\left[\tilde{\mathnormal{z}}^{\textup{adv}}_1\right]=\mathnormal{y},\quad
\mathop{\mathbb{E}}\left[\tilde{\mathnormal{z}}^{\textup{adv}}_i\right]=
(\eta-\lambda)\mathnormal{y}.
\end{equation}
\end{lemma}
Proof is in Appendix~\ref{append:proofs}.
The stochastic gradient descent to the cross-entropy loss on $\tilde{\bm{\mathnormal{z}}}^{\textup{adv}}$ is applied to update our classification model.
In particular, by deriving the auxiliary loss gradient with respect to the adversarial feature vector, we determine the training signals that are generated from the auxiliary task and transferred to the primary task through the shared feature embedding function.
\newtheorem{theorem}{Theorem}
\begin{theorem}\label{theorem1}
Let $\ell(;\bm{\mathnormal{w}})$ and $\tilde{\ell}(;\gamma\bm{\mathnormal{w}})$ be the loss functions of the primary and auxiliary tasks, respectively, and $t=\frac{1}{2}(y+1)$. When the auxiliary data are closely related to the primary data from the perspective of robust and non-robust features, i.e., $\abs{\gamma}=1$, the expectation of the gradient of $\tilde{\ell}$ with respect to $\tilde{\mathnormal{z}}^{\textup{adv}}_i:i\in\{2,\cdots,d+1\}$ is
\begin{equation}
\begin{gathered}
    \mathbb{E}\left[\frac{\partial\tilde{\ell}}{\partial\tilde{\mathnormal{z}}^{\textup{adv}}_i}\right]=
    \frac{1}{d}\mathbb{E}\left[\sigma(\bm{\mathnormal{w}}^\top\bm{\mathnormal{z}}^{\textup{adv}})-\mathnormal{t}\right]=
    \mathbb{E}\left[\frac{\partial\ell}{\partial{\mathnormal{z}}^{\textup{adv}}_i}\right].
\end{gathered}
\end{equation}
\end{theorem}
The theoretical results in the cases of $\abs{\gamma}<1$ (weak correlation) are discussed in Appendix~\ref{append:proofs}.
From Lemma~\ref{lemma2} and Theorem~\ref{theorem1}, for $i\in\{2,\cdots,d+1\}$, it can be seen that $\textup{sign}\left(\mathop{\mathbb{E}}\left[\tilde{\mathnormal{z}}^{\textup{adv}}_i\right]\right)=\textup{sign}\left(\mathbb{E}\left[\frac{\partial\tilde{\ell}}{\partial\tilde{\mathnormal{z}}^{\textup{adv}}_i}\right]\right)$.
That is, the application of a gradient descent guides the shared feature embedding function to pay less attention to non-robust features. In addition, Theorem~\ref{theorem1} shows that if the auxiliary task is closely related to the primary task in terms of non-robust features, the training signals obtained from the auxiliary adversarial loss and back-propagated to the shared feature embedding function have the same effect as those of the primary task from the perspective of non-robust feature regularization. Therefore, this can be considered data amplification for non-robust feature regularization.

\paragraph{Robust feature learning.}\label{sec:robust feature transfer}
If $\abs{\gamma}=1$ and the weight value for the robust feature $\mathnormal{z}_1$ is non-zero, clearly, the auxiliary task on $\tilde{\bm{\mathnormal{z}}}^{\textup{adv}}$ can induce data amplification for robust feature learning as well as non-robust feature regularization.
However, when the auxiliary dataset contains extraneous robust features, the learning on $\tilde{\bm{\mathnormal{z}}}^{\textup{adv}}$ may lead to negative transfer, which suppresses the advantages of multi-domain learning.
To prevent inductive transfer between tasks, Caruana \cite{caruana1997multitask} shuffled the class labels among all samples in the auxiliary dataset. Similarly, to avoid negative transfer, the use of random labels can be considered in our case.
To investigate the effect of adversarial training on the shuffled auxiliary dataset, we replace the true labels in the auxiliary data defined in Eq.~(\ref{datamodel}) with random labels.
That is, we define the auxiliary feature--label pairs, $(\tilde{\bm{\mathnormal{z}}},\mathnormal{q})\in\mathbb{R}^{d+1}\times\{\pm1\}$, sampled from a distribution as follows:
\begin{equation}\label{shuffled data}
\begin{gathered}
    \tilde{\mathnormal{z}}_1\stackrel{}{\sim}\mathcal{N}(y\abs{\gamma},\mathnormal{v}^2),\;
    \tilde{\mathnormal{z}}_2,\dots,\tilde{\mathnormal{z}}_{d+1}\stackrel{i.i.d.}{\sim}\mathcal{N}(\eta\mathnormal{y}\abs{\gamma}, 1),\\
    \mathnormal{q}\stackrel{u.a.r}{\sim} \{-1, +1\}.
\end{gathered}
\end{equation}
For the case in which the auxiliary task is adversarially trained on $(\tilde{\bm{\mathnormal{z}}},\mathnormal{q})$ pairs with the cross-entropy loss, the following theorem can be proven:
\begin{theorem} \label{theorem2}
Let $\tilde{\ell}\left(;\gamma\bm{\mathnormal{w}}\right)$ be the loss function of the auxiliary task. Then, if $\abs{\gamma}=1$, with high probability, the signs of $\tilde{\mathnormal{z}}^{\textup{adv}}_i:i\in\{2,\cdots,d+1\}$ and the auxiliary loss gradient with respect to $\tilde{\mathnormal{z}}^{\textup{adv}}_i$ are
\begin{equation}
    \textup{sign}\left(\tilde{\mathnormal{z}}^{\textup{adv}}_i\right)
    =-\gamma\mathnormal{q}
    =\textup{sign}\left(\frac{\partial\tilde{\ell}}{\partial\tilde{\mathnormal{z}}^{\textup{adv}}_i}\right).
\end{equation}
\end{theorem}
Because the gradient with respect to $\tilde{\mathnormal{z}}^{\textup{adv}}_i$ is of the same sign as $\tilde{\mathnormal{z}}^{\textup{adv}}_i$ with high probability, the application of a gradient descent makes the shared feature embedding function to refrain from using non-robust features, thereby enabling the model to achieve non-robust feature regularization.
Conversely, the shuffled dataset cannot provide any robust features because the use of random labels completely eliminates the relationship between images and labels.
To further investigate the effect of adversarial training on the shuffled auxiliary dataset with regard to robust feature learning, we assign a positive number to the weight (defined in Eq.~(\ref{our_classifier})) corresponding to the robust feature $\mathnormal{z}_1$ and derive the training signals that are sent to the shared feature embedding function as follows:
\begin{theorem} \label{theorem3}
Let $\tilde{\ell}(;\gamma\bm{\mathnormal{w}})$ be the loss function of the auxiliary task. Then, if $\abs{\gamma}=1$ and $\mathnormal{w}_1>0$, with high probability, the signs of $\tilde{\mathnormal{z}}^{\textup{adv}}_1$ and the auxiliary loss gradient with respect to $\tilde{\mathnormal{z}}^{\textup{adv}}_1$ are
\begin{equation}
    \textup{sign}(\tilde{\mathnormal{z}}^{\textup{adv}}_1)
    =\mathnormal{y},\quad
    \textup{sign}\left(\frac{\partial\tilde{\ell}}{\partial\tilde{\mathnormal{z}}^{\textup{adv}}_1}\right)
    =-\gamma\mathnormal{q}.
\end{equation}
\end{theorem}
Assuming that the classification model is still vulnerable to adversarial examples, $\abs{\frac{\partial\tilde{\ell}}{\partial\tilde{\mathnormal{z}}^{\textup{adv}}_i}}$ is independent of $q$ because an adversary can always yield a large loss regardless of $q$. Hence, in the theoretical case of an infinite batch size, the adversarial training on the shuffled auxiliary dataset will not affect the robust feature learning for the primary task because $\mathnormal{y}$ and $\mathnormal{q}$ are independent of each other, and $\mathnormal{q}$ is sampled uniformly at random.
In practice, however, the minibatch gradient descent is employed to train DNNs, and thus, unfavorable training signals can be generated from the auxiliary task on the shuffled dataset in terms of robust feature learning.
To resolve this issue, we use the expectation of random labels instead of the one-hot random labels. 
Furthermore, to close the gap between the theory~($\abs{\gamma}=1$) and practice, we use a shared prediction function for the primary and the shuffled auxiliary data.
In other words, we assign $\mathnormal{y}^{\textup{ER}}=[\frac{1}{c},\dots,\frac{1}{c}]$, where $\mathnormal{c}$ is the number of the classes in the primary dataset, to the auxiliary data to observe only the effectiveness of non-robust feature regularization while excluding the contribution of robust feature learning.
\subsection{Empirical evidence}\label{sec:empirical}
\input{Table/table1_suggested} 
To empirically demonstrate the arguments developed above, we conduct a test confirming the following two statements: (\romannumeral 1)~When $\tilde{D}$ has a weak relationship with $D$ in terms of robust features, the use of $y^{\textup{ER}}$ ($I_\textup{high}=\emptyset$ and $I_\textup{low}=\{0,1,\cdots,|\tilde{D}|-1\}$ in Fig.~\ref{fig_main}) results in better adversarial robustness than that produced by the use of $\tilde{y}$ ($I_\textup{high}=\{0,1,\cdots,|\tilde{D}|-1\}$ and $I_\textup{low}=\emptyset$ in Fig.~\ref{fig_main}), and (\romannumeral 2)~when $\tilde{D}$ is closely related to $D$ in terms of robust features, the use of $y^{\textup{ER}}$ results in worse adversarial robustness than that produced by the use of $\tilde{y}$.
Table~\ref{tab:table-1} lists the results of executing the test using CIFAR-10 as the primary dataset $D$. Here, we use SVHN \cite{svhn_dataset}, CIFAR-100, and Places365 as auxiliary datasets that are weakly related to CIFAR-10 in terms of robust features, based on the previous OOD detection studies \cite{outlier,ssd}. In addition, we use ImageNet as an auxiliary dataset that is closely related to CIFAR-10 from the perspective of robust features \cite{pretrain_hendrycks,simclr}.
As shown, for SVHN, CIFAR-100, and Places365, the use of $y^{\textup{ER}}$ leads to better adversarial robustness than that induced by the use of $\tilde{y}$, even though image-label mappings in the auxiliary datasets are disrupted.
These results demonstrate that the robust feature learning induced by using all the image--label pairs ($\tilde{x}$--$\tilde{y}$) in each of the SVHN, CIFAR-100, and Places365 datasets is detrimental to the primary task on CIFAR-10.
By contrast, for ImageNet, the fact that blocking robust feature learning using $y^{\textup{ER}}$ leads to less performance improvement indicates that beneficial inductive transfer in terms of robust feature learning can be achieved from the auxiliary task on ImageNet. That is, ImageNet shares a large number of robust features as well as non-robust features with CIFAR-10.
\subsection{A confidence-based selection strategy}\label{Methods: confidence-based approach}
\input{Figure/algorithm_suggested}
Our analysis demonstrate that when using an auxiliary dataset for the primary task, the optimal algorithm to be applied varies according to the relationship between the two datasets in terms of robust features. In real world scenario, however, the auxiliary dataset may contain both favorable and unfavorable robust features for the primary task. Hence, we introduce a sample-wise selection strategy in our proposed method. The selection strategy is required to classify given auxiliary data into two groups based on the robust features, and the robust features exhibit human-perceptible patterns as shown in Fig.~\ref{fig_rob_nonrob}. Therefore, the objective of the selection strategy is consistent with that of existing OOD detection methods \cite{outlier,ssd}.
Hendrycks et al. \cite{outlier} proposed an OOD detection method using the confidence scores of the query data. To be specific, they trained models to give OOD samples a uniform posterior. Interestingly, the use of $y^{\textup{ER}}$ is naturally connected to their proposed method. That is, the use of $y^{\textup{ER}}$ achieves non-robust feature regularization without resulting in negative transfer while at the same time lowering the confidence scores of data irrelevant to the primary task in terms of robust features. On this basis, our selection strategy classify given auxiliary data samples based on their confidence scores.
The proposed confidence-based method first (\romannumeral 1) trains a classifier from scratch on the primary dataset; (\romannumeral 2) after a few epochs (warm-up), sets up a threshold using a hyperparamter $\pi\in\mathbb{R}^+$ and the mean confidence of the sampled primary data to sort out the auxiliary data samples that are likely to cause negative transfer; (\romannumeral 3) selects the lower-than-threshold auxiliary data in each training batch based on their confidence scores for the primary classes; (\romannumeral 4) uses the low confidence data with $y^{\textup{ER}}$ and the remaining auxiliary data with $\tilde{y}$.
The pseudo-code is provided in Algorithm~\ref{algo}. 

%% file: Figure/fig_main.tex
\begin{figure}[t]
\begin{center}
\includegraphics[width=1.0\linewidth]{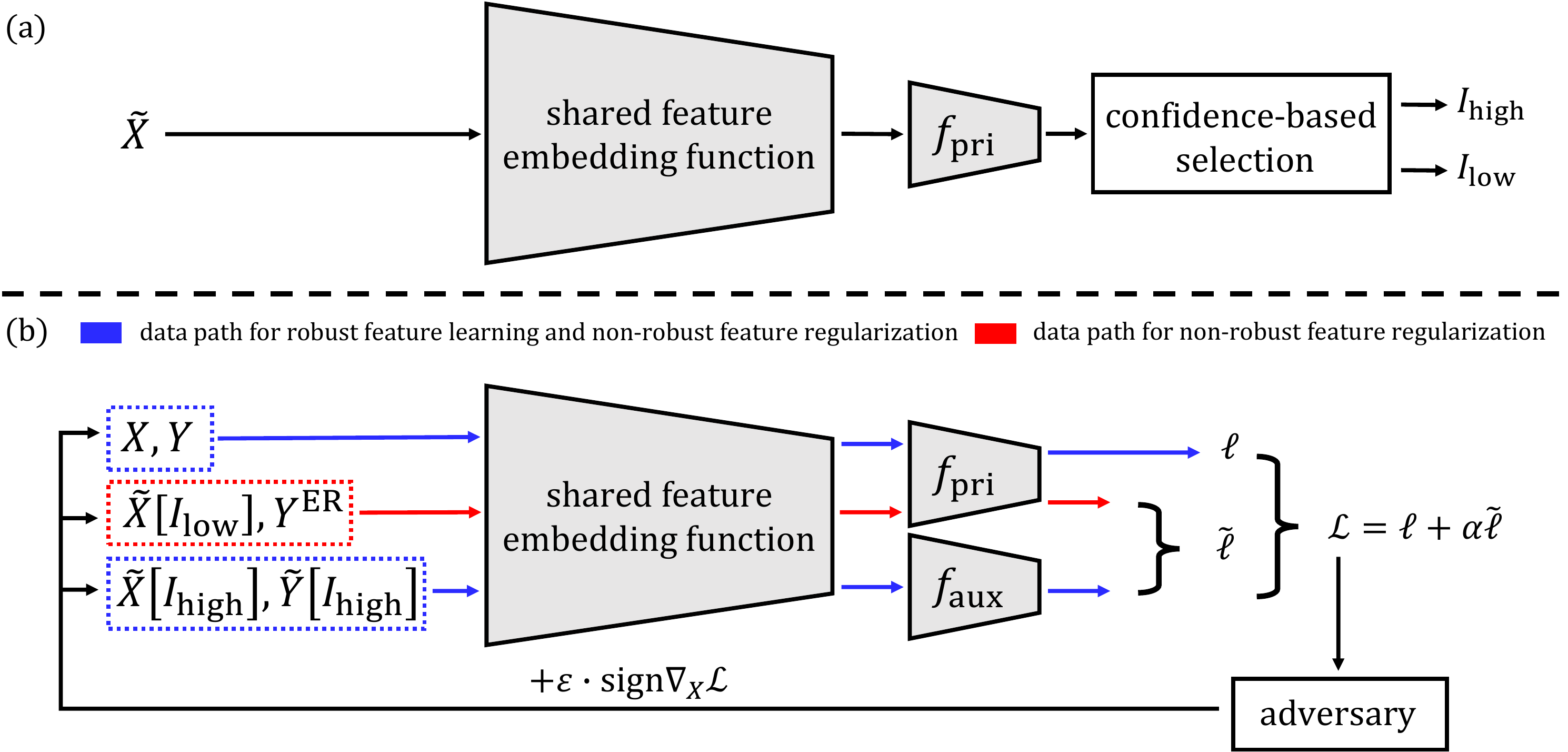}
\end{center}
   \caption{{\textbf{Overview of BiaMAT.} We use a shared feature embedding function for both primary and auxiliary tasks, while prediction functions ($f_\textup{pri}$ and $f_\textup{aux}$) are separated for each task. (a) At the first stage of BiaMAT, a confidence-based selection strategy classify given auxiliary data $\tilde{X}$ into two groups: auxiliary data that share robust and non-robust features with the primary dataset ($\tilde{X}[I_\text{high}]$); and that share only non-robust features with the primary dataset ($\tilde{X}[I_\text{low}]$). (b) We adversarially train the model on $\tilde{X}[I_\text{high}]$ and the primary data $X$ using their ground truth labels ($\tilde{Y}[I_\text{high}]$ and $Y$) to achieve robust feature learning and non-robust feature regularization (indicated in \textcolor{blue}{blue}). For $\tilde{X}[I_\text{low}]$, however, we use the expectation of random labels $y^{\textup{ER}}$ to induce data amplification only for non-robust feature regularization (indicated in \textcolor{red}{red})}}
\label{fig_main}
\end{figure}

%% file: Table/table1_suggested.tex
\newcommand{\rednum}[1]{\textcolor{red}{#1}}

\begin{table}[]
\centering
\caption{Accuracy (under Autoattack \cite{auto_attack}) comparison of the models adversarially trained \cite{pgd_attack} on CIFAR-10 using either $\tilde{y}$ or $y^{\textup{ER}}$ for the auxiliary dataset. The best result for each auxiliary dataset is indicated in bold; auxiliary datasets that produce better results when used with $y^{\textup{ER}}$ are indicated in \rednum{red}}\label{tab:table-1}
\vskip 5mm

\footnotesize
\addtolength{\tabcolsep}{-2pt}
\begin{tabular}{cccccc}

\toprule
\multirow{2}{*}{$\tilde{y}/y^{\textup{ER}}$} & \multicolumn{4}{c}{Auxiliary dataset} & \multirow{2}{*}{Baseline \cite{pgd_attack}}  \\
& \rednum{SVHN} &\rednum{CIFAR-100} & \rednum{Places365} & ImageNet &         \\ \midrule[.1em]
$\tilde{y}$   & 47.44 & 48.48 & 48.88 & \textbf{50.33}& \multirow{2}{*}{48.53}\\
$y^{\textup{ER}}$& \textbf{48.53} & \textbf{49.89} & \textbf{49.24} & 49.81&\\\midrule[.1em]
\end{tabular}
\end{table}

%% file: Figure/algorithm_suggested.tex
\begin{figure}[t]
{
\hfill
\begin{minipage}{1.0\linewidth}
  \begin{algorithm}[H]
\caption{Biased multi-domain adversarial training (BiaMAT) with the confidence-based selection strategy}
\begin{algorithmic}[1]
\label{algo}
\REQUIRE Primary dataset $D$, auxiliary dataset $\tilde{D}$, model parameter $\bm{\theta}$, training iterations $K$, warmup iterations $K_w$, learning rate $\tau$, hyperparameters $\alpha\in\mathbb{R}^{+}$ and $\pi\in\mathbb{R}^{+}$
\FOR{$k=1$ \textbf{to} $K_w$}
\STATE \textcolor{blue}{/* Warm-up training on $D$ */}
\STATE Sample a minibatch $B\mathtt{\sim}D$
\STATE $\mathcal{L} \leftarrow \mathop{\mathbb{E}}_{(\bm{\mathnormal{x}},\mathnormal{y})\mathtt{\sim}B} \left[\ell_{\textup{adv}}(\bm{\mathnormal{x}},\mathnormal{y};h_\textup{pri},S)\right]$
\STATE $\bm{\theta} \leftarrow \bm{\theta} - \tau\cdot\nabla_{\theta}\mathcal{L}$
\ENDFOR
\STATE  \textcolor{blue}{/* Confidence threshold */}
\STATE $\omega \leftarrow \pi\cdot\mathop{\mathbb{E}}_{(\bm{\mathnormal{x}},\mathnormal{y})\mathtt{\sim}B} \left[\max h_{\textup{pri}}(\bm{\mathnormal{x}})\right]$
\FOR{$k=K_w + 1$ \textbf{to} $K$}
\STATE Sample a minibatch pair $B\mathtt{\sim}D$ and $\tilde{B}\mathtt{\sim}\tilde{D}$
\STATE $\ell \leftarrow \mathop{\mathbb{E}}_{(\bm{\mathnormal{x}},\mathnormal{y})\mathtt{\sim}B} \left[\ell_{\textup{adv}}(\bm{\mathnormal{x}},\mathnormal{y};h_\textup{pri},S)\right]$
\STATE $\ell_{\textup{high}}, \ell_{\textup{low}} \leftarrow 0, 0$
\STATE \textcolor{blue}{/* Confidence-based selection strategy */}
\FOR{$\tilde{\bm{\mathnormal{x}}}, \tilde{\mathnormal{y}}$ \textbf{in} $\tilde{B}$}
\IF{$\max h_{\textup{pri}}(\tilde{\bm{\mathnormal{x}}})<\omega$}
\STATE $\ell_{\textup{low}}\mathrel{{+}{=}} \frac{1}{|\tilde{B}|}\ell_{\textup{adv}}(\tilde{\bm{\mathnormal{x}}},\mathnormal{y}^{\textup{ER}};h_{\textup{pri}},S)$
\ELSE
\STATE $\ell_{\textup{high}}\mathrel{{+}{=}} \frac{1}{|\tilde{B}|}\ell_{\textup{adv}}(\tilde{\bm{\mathnormal{x}}},\tilde{\mathnormal{y}};h_{\textup{aux}},S)$
\ENDIF
\ENDFOR
\STATE $\mathcal{L} \leftarrow \ell+\alpha(\ell_{\textup{low}} + \ell_{\textup{high}})$
\STATE $\bm{\theta} \leftarrow \bm{\theta} - \tau\cdot\nabla_{\theta}\mathcal{L}$
\ENDFOR
\STATE \textbf{Output:} adversarially robust classifier $h_{\textup{pri}}=f_{\textup{pri}}\circ g$
\end{algorithmic}
\end{algorithm}
\end{minipage}
\par
}
\end{figure}

%% file: exp.tex
\section{Experimental results and discussion} \label{sec:experiments}
\subsection{Experimental setup}\label{sec:expsetup}
\paragraph{Datasets.}
We complement our analysis with experiments conducted on the CIFAR datasets and ImageNet. ImageNet is resized~\cite{downsampled_imagenet} to dimensions of $64\times64$ and then randomly divided into datasets that contain 100 and 900 classes, which are termed ImgNet100 and ImgNet900, respectively. SVHN~\cite{svhn_dataset}, Places365, and ImageNet are used as auxiliary datasets. Auxiliary data that do not fit the input size of the classifier are resized to the primary data size. For instance, when CIFAR-10 is used as the primary dataset, Places365 is down-sampled to a dimension of $32\times32$, and ImageNet32x32 is leveraged.

\paragraph{Adversarial attack methods.}
Fast gradient sign method (FGSM)~\cite{fgsm_attack} is an one-step attack using the sign of the gradient. Madry et al.~\cite{pgd_attack} proposed an iterative application of the FGSM method (PGD). Carlini \& Wagner (CW)~\cite{cw_attack} attack is a targeted attack that maximize the logit of a target class and minimize that of ground-truth. Autoattack (AA)~\cite{auto_attack} is an ensemble attack that consists of two PGD extensions, one white-box attack~\cite{fab_attack}, and one black-box attack~\cite{square_attack}. We focus on the $\ell_\infty\text-$robustness, the most common robustness scenario considered in the field of heuristic defenses~\cite{pgd_attack,trades_defense,avmixup}.
\paragraph{Implementation details.}\label{sec:implementation details}
In our experiments, we adopt the adversarial training methods proposed by Madry et al.~\cite{pgd_attack} and Zhang et al.~\cite{trades_defense} as the baseline methods, denoted by AT and TRADES, respectively. On the CIFAR datasets, we use WRN28-10~\cite{wide_resnet} and WRN34-10 for AT and TRADES, respectively.
Although increasing the number of training epochs is expected to lead to higher adversarial robustness because of the use of additional data, owing to the high-computational complexity of adversarial training, we restrict the training of BiaMAT to 100 or 110 epochs with a batch size of 256~(128 primary and 128 auxiliary data samples, respectively).
To evaluate adversarial robustness, we apply multiple attacks, including PGD, CW, and AA, with an $\ell_\infty\text-$bound with the same setting as that used in the training. PGD and CW with $K$~iterations are denoted by PGD\textsuperscript{$K$} and CW\textsuperscript{$K$}, respectively, and the unperturbed test set is denoted by Clean. We consistently select the best checkpoint~\cite{overfitting_adv} to measure the adversarial robustness of the model on the test set.
Further details regarding the model implementation, including an ablation study on choosing different values of $\pi$, are summarized and discussed in Appendix~\ref{append-implementation}.
\subsection{Adversarial robustness under various attacks}
\input{Figure/fig_ratio}
\input{Table/table2}
Table \ref{tab:table-2} summarizes the improvements in the adversarial robustness of the models obtained from the application of BiaMAT. The proposed method can freely use various auxiliary datasets as it avoids negative transfer through the application of the confidence-based selection strategy; in fact, a comparison of Tab.~\ref{tab:table-1} and Tab.~\ref{tab:table-2} demonstrates that the proposed method effectively overcomes negative transfer and achieves only beneficial training signals for the primary task from the auxiliary task.
To observe how the confidence-based selection strategy works while the model is being trained through the application of the proposed method, we define a ratio $\frac{n_\textup{high}}{n_\textup{aux}}$, where $n_\textup{aux}$ and $n_\textup{high}$ denote the amount of auxiliary data and the higher-than-threshold auxiliary data within each training batch, respectively. That is, the ratio represents the percentage of data used for robust feature learning as well as non-robust feature regularization for an auxiliary dataset.
Figure~\ref{fig_ratio} shows the plot of the ratio $\frac{n_\textup{high}}{n_\textup{aux}}$ at $\pi=0.55$ (defined in Algorithm~\ref{algo}) during the training of the AT+BiaMAT models using various auxiliary datasets on CIFAR-10. As shown, the confidence-based selection strategy successfully filters out data that are likely to induce negative transfer for the primary task. In other words, a relatively high percentage of ImageNet data are used for robust feature learning, and each of the SVHN, CIFAR100, and Places365 datasets are mostly used with $y^{\textup{ER}}$, which is consistent with the results listed in Tab.~\ref{tab:table-1}. Additional analysis is in Appendix~\ref{append:analysisconfidence}.

\input{Table/table_comp}
We conduct several experiments to further investigate the proposed method. (Appendix~\ref{append:moreaux})~To observe the effects of the use of more auxiliary datasets, we train a BiaMAT model using a combination of two auxiliary datasets; 
the results show that the use of more auxiliary datasets does not always lead to further improvements in adversarial robustness. 
In other words, the relationship (in terms of robust and non-robust features) between the primary and auxiliary datasets is more important to BiaMAT than the number of auxiliary datasets.
(Appendix~\ref{append:robustdataset})~To further demonstrate that robust feature learning can be achieved from the auxiliary task in BiaMAT, we construct robust datasets~\cite{not_bugs_are_features} from the AT and AT+BiaMAT models and normally train models from scratch on each robust dataset ($D^\textup{AT}$ and $D^\textup{BiaMAT}$);
the results show that $D^\textup{BiaMAT}$ results in more robust models than those trained on $D^\textup{AT}$, implying that BiaMAT enables DNNs to learn better robust features via inductive transfer between adversarial
training on the primary and auxiliary datasets.

\subsection{Comparison with other related methods}\label{subsec:comp}
Carmon et al.~\cite{unlabeled_adv} proposed a semi-supervised learning technique where the training dataset is augmented with unlabeled in-distribution data;
the main difference between this and BiaMAT is the distribution of additional data. For instance, Carmon et al. collected the in-distribution data of CIFAR-10 from 80 Million Tinyimages dataset~\cite{80mti} and used the unlabeled data with pseudo-labels.
Therefore, no assumptions are required regarding the classes of the primary and auxiliary datasets in our scenario, but the semi-supervised method is ineffective when the primary and auxiliary datasets do not share the same class distributions.
To demonstrate this, we assign pseudo-labels to the auxiliary data using a pre-trained classifier and configure each training batch (for TRADES) such that it contains the same amount of primary and pseudo-labeled data, as in \cite{unlabeled_adv}.
In particular, we sort ImageNet based on the confidence in the CIFAR-10 classes and select the top 50k (or top 5k) samples for each class in CIFAR-10 (or CIFAR-100); this is denoted as ImageNet-500k.
As shown in Tab.~\ref{tab:table-comp}, the Carmon et al.~\cite{unlabeled_adv} method exhibits lower effectiveness than the proposed method. Specifically, the results obtained using CIFAR-100 and Places365 demonstrate that the semi-supervised method is vulnerable to negative transfer because of the considerable domain discrepancy between the primary and auxiliary datasets.

Hendrycks et al.~\cite{pretrain_hendrycks} demonstrated that ImageNet pre-training can improve adversarial robustness on the CIFAR datasets. However, the pre-training method is effective only when a dataset that has a distribution similar to that of the primary data and a sufficiently large number of samples is used. To demonstrate this, we adversarially pre-train the CIFAR-100 and ImageNet~\cite{pretrain_hendrycks} models and then adversarially fine-tune them on CIFAR-10. The results in Tab.~\ref{tab:table-comp} demonstrate that the pre-training method is ineffective when leveraging datasets that do not satisfy the conditions mentioned above. In other words, because the effect achieved by the pre-training method arises from the reuse of features pre-trained on a dataset that contains a large quantity of data with a distribution similar to that of the primary dataset, CIFAR-100 is not suitable for application of the CIFAR-10 task. Conversely, BiaMAT avoids such negative transfer through the application of a confidence-based strategy. That is, these results emphasize the high compatibility of the proposed method with a variety of datasets.
Additional experimental results, including comparisons with OAT \cite{oat} or a generative model-based method \cite{advddpm}, can be found in Appendix~\ref{append:comp}.

%% file: Figure/fig_ratio.tex
\begin{figure}[t]
\begin{center}
\raisebox{0pt}[\dimexpr\height-0.0\baselineskip\relax]{\includegraphics[width=0.9\linewidth]{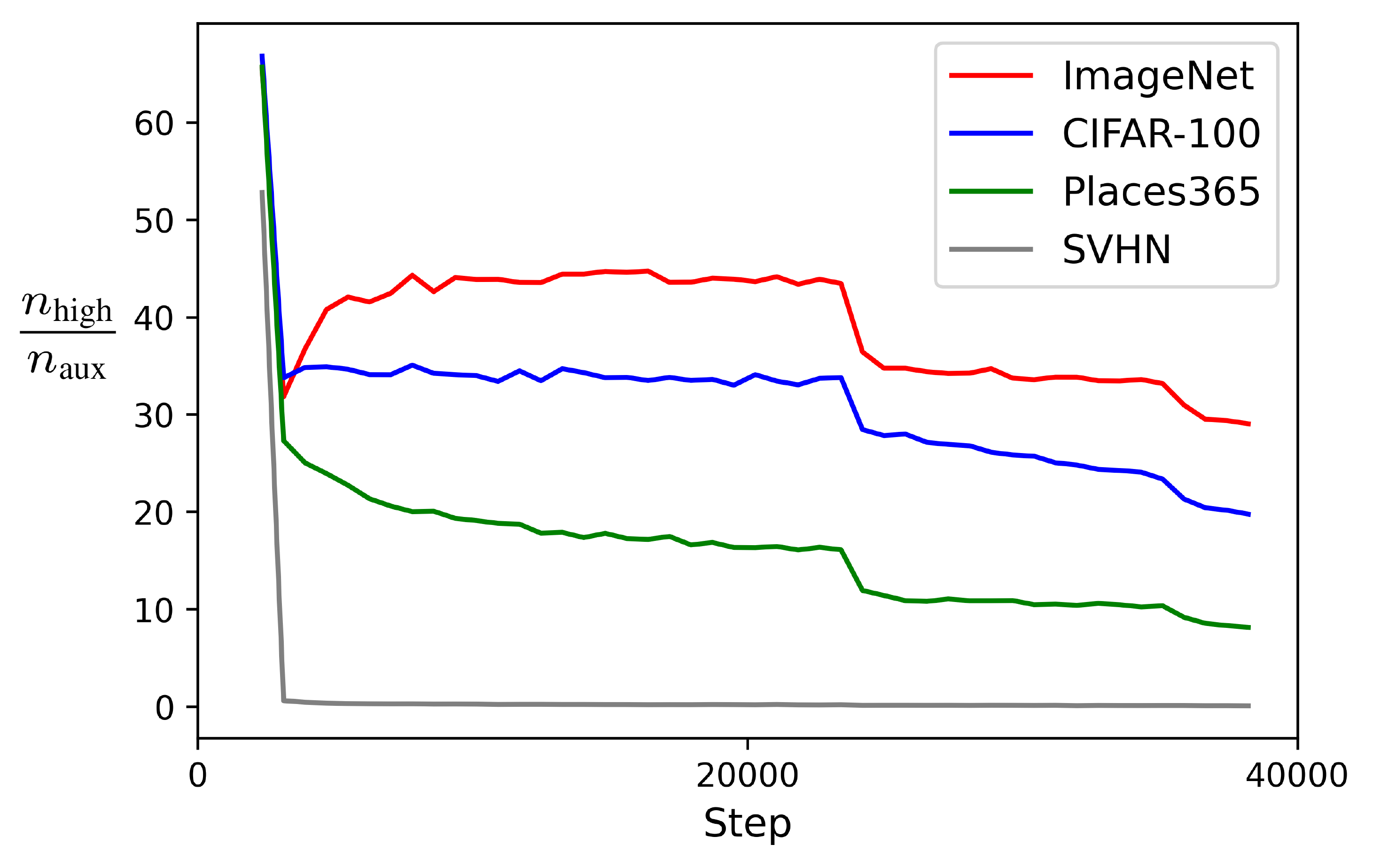}}
\end{center}
\vspace{-10pt}
\caption{Ratio $\frac{n_{\textup{high}}}{n_{\textup{aux}}}$ for each auxiliary dataset with respect to the primary task on CIFAR-10.}
\label{fig_ratio}
\end{figure}

%% file: Table/table2.tex
\begin{table*}[t]
\centering
\caption{Performance improvements~(accuracy \%) on CIFAR-10, CIFAR-100, and ImgNet100 following application of the proposed method using various datasets. The best results within each baseline method~(AT and TRADES) are indicated in bold}
\begin{tabular*}{\textwidth}{c @{\extracolsep{\fill}} cccccc}
\toprule
Primary dataset                  & Method                   & Auxiliary dataset        & Clean         & PGD\textsuperscript{$100$}          & CW\textsuperscript{$100$}    & AA         \\ \midrule[.1em]
\multirow{11}{*}{CIFAR-10}  
                          & \multirow{1}{*}{AT}   & -    & 87.37 & 50.87 & 50.93 & 48.53 \\\cmidrule{2-7}
                          & \multirow{4}{*}{\begin{tabular}[c]{@{}c@{}}AT+BiaMAT\\ (ours)\end{tabular}} & SVHN & 87.34 & 51.90 & 51.40 & 48.61 \\
                          && CIFAR-100       & 87.22 & 55.93 & 52.09 & 50.08 \\
                          &  & Places365     & 87.76 & 57.00 & 51.70 & 49.48 \\
                          &  & ImageNet      & \textbf{88.75} & \textbf{57.63} & \textbf{53.04} & \textbf{50.78} \\\cmidrule{2-7}
                          & \multirow{1}{*}{TRADES}          & -& 85.85 & 56.62 & 55.16 & 53.93 \\\cmidrule{2-7}
                          & \multirow{4}{*}{\begin{tabular}[c]{@{}c@{}}TRADES+BiaMAT\\ (ours)\end{tabular}} 
                          & SVHN     & 85.49 & 56.86 & 55.21 & 53.94 \\
                          & & CIFAR-100     & 87.02 & 58.69 & 56.85 & 55.48 \\
                          & & Places365     & 87.18 & 59.15 & 56.36 & 55.24 \\
                          & & ImageNet & \textbf{88.03} & \textbf{59.80} & \textbf{58.01} & \textbf{56.64}
                           \\\midrule[.1em]
                           \multirow{7}{*}{CIFAR-100} 
                          & \multirow{1}{*}{AT}  &-  & 62.59 & 26.80 & 26.07 & 24.13 \\\cmidrule{2-7}
                          & \multirow{2}{*}{\begin{tabular}[c]{@{}c@{}}AT+BiaMAT\\ (ours)\end{tabular}} &Places365 & 63.44 & 32.61 & 28.53 & 26.49 \\
                          && ImageNet  & \textbf{64.05} & \textbf{33.74} & \textbf{29.78} & \textbf{27.65} \\\cmidrule{2-7}
                          & \multirow{1}{*}{TRADES} & -  & 62.04 & 32.53 & 30.07 & 28.82 \\\cmidrule{2-7}
                          & \multirow{2}{*}{\begin{tabular}[c]{@{}c@{}}TRADES+BiaMAT\\ (ours)\end{tabular}} & Places365     & 64.58 & 34.38 & 30.72 & 29.24 \\
                          & & ImageNet & \textbf{65.82} & \textbf{36.36} & \textbf{33.42} & \textbf{31.87} \\\midrule[.1em]
\multirow{7}{*}{ImgNet100}     
                          & \multirow{1}{*}{AT}  & -  & 66.60 & 35.46 & 31.90 & 29.54 \\\cmidrule{2-7}
                          & \multirow{2}{*}{\begin{tabular}[c]{@{}c@{}}AT+BiaMAT\\ (ours)\end{tabular}} & Places365 & \textbf{70.04} & \textbf{40.52} & 33.24 & 30.64 \\
                          && ImgNet900           & 68.00 & 40.18 & \textbf{35.00}     & \textbf{32.88} \\ \cmidrule{2-7}
                          & \multirow{1}{*}{TRADES} & -  & 56.16 & 27.90 & 22.98 & 21.90 \\\cmidrule{2-7}
                          & \multirow{2}{*}{\begin{tabular}[c]{@{}c@{}}TRADES+BiaMAT\\ (ours)\end{tabular}} & Places365     & 57.80 & 29.30 & 24.14 & 23.06 \\
                          & & ImageNet & \textbf{58.76} & \textbf{31.26} & \textbf{25.98} & \textbf{24.98} \\\midrule[.1em]
 
\end{tabular*}
\label{tab:table-2}
\end{table*}

%% file: Table/table_comp.tex
\begin{table*}[]
\centering
\caption{Comparison (accuracy \%) of the effectiveness of BiaMAT with the semi-supervised~\cite{unlabeled_adv} and pre-training~\cite{pretrain_hendrycks} methods on CIFAR-10.}
\begin{tabular*}{\textwidth}{c @{\extracolsep{\fill}} ccccc}
\toprule
Method & Auxiliary dataset & Clean & PGD\textsuperscript{$100$} & CW\textsuperscript{$100$} & AA \\ \midrule[.1em]
\multirow{1}{*}{TRADES (baseline)}   & -   & 85.85 & 56.62 &  55.16 & 53.93 \\\midrule[.05em]
\multirow{2}{*}{Hendrycks \etal~\cite{pretrain_hendrycks}}   & CIFAR-100    & 80.21 & 45.68 & 44.52 & 42.36 \\
& ImageNet     & 87.11 & 57.16 & 55.43 & 55.30\\\midrule[.05em]
\multirow{4}{*}{Carmon \etal~\cite{unlabeled_adv}}
& CIFAR-100     & 82.61 & 54.32 & 51.64  & 50.81 \\
& Places365     & 83.95 & 56.72 & 53.95 & 52.81 \\
& ImageNet     & 85.42 & 57.46 & 54.66 &  53.79 \\
& ImageNet-500k & 86.02 & 59.49 & 56.43 & 55.63\\\midrule[.05em]
\multirow{3}{*}{\begin{tabular}[c]{@{}c@{}}TRADES+BiaMAT\\ (ours)\end{tabular}} 
& CIFAR-100     & \textbf{87.02} & \textbf{58.69} & \textbf{56.85} & \textbf{55.48} \\
& Places365     & \textbf{87.18} & \textbf{59.15} & \textbf{56.36} & \textbf{55.24} \\
& ImageNet     & \textbf{88.03} & \textbf{59.80} & \textbf{58.01} &  \textbf{56.64} \\\midrule[.1em]
\end{tabular*}
\label{tab:table-comp}
\end{table*}

%% file: conclusion.tex
\section{Conclusions and future directions}\label{sec:conclusion}

In this study, we develop BiaMAT, a method that uses publicly available (labeled) auxiliary datasets to reduce the large gap between training and test errors in adversarial training.
Our theoretical and empirical analysis demonstrate that the effectiveness of BiaMAT can be attributed to two factors: non-robust feature regularization and robust feature learning.
In particular, we show that while existing methods are vulnerable to negative transfer due to the distributional discrepancy between auxiliary and primary data, BiaMAT can successfully overcome negative transfer through the application of a confidence-based selection strategy.
In this study, however, the application of any method that can improve the performance of multi-domain learning is not considered. In addition, there is room for improvement in the effectiveness of BiaMAT with regard to the strategy used to avoid negative transfer.
In future work, therefore, we will develop algorithms in which additional techniques, such as the use of adaptive weighting strategies~\cite{adaptive_multitask}, are implemented.

%% file: appendix.tex

\section{Proofs}\label{append:proofs}
\setcounter{equation}{7}
\newtheorem{lemma-append}{Lemma}
\setcounter{lemma-append}{1}
\begin{lemma-append}\label{lemma-append}
Let $i\in\{2,\cdots,d+1\}$ and $\eta<\lambda<1$. Then, the expectation of the adversarial feature vector against the auxiliary task is
\begin{equation}\label{eq:lemma-append}
\mathop{\mathbb{E}}\left[\tilde{\mathnormal{z}}^{\textup{adv}}_1\right]=\mathnormal{y},\quad
\mathop{\mathbb{E}}\left[\tilde{\mathnormal{z}}^{\textup{adv}}_i\right]=
(\eta-\lambda)\mathnormal{y}.
\end{equation}
\end{lemma-append}
\setcounter{equation}{12}
\begin{proof}
Our model comprises a non-linear feature embedding function $g:\mathcal{X}\rightarrow\mathcal{Z}$ and a linear classifier $f_{\gamma\bm{\mathnormal{w}}}:\mathcal{Z}\rightarrow\mathcal{Y}$. In addition, the theoretical model is based on two principles that reflect the behaviors of neural networks against adversarial examples: (\romannumeral 1)~the signs of the non-robust features $\tilde{z}_i:i\in\{2,\cdots,d+1\}$ are switched by an adversary with high probability; (\romannumeral 2)~the sign of the robust feature
$z_1$ is not easily switched by an adversary. The objective of an adversary is to find an adversarial perturbation $\bm{\delta}^\star=\argmax_{\bm{\delta}\in S}\tilde{\ell}(g(\tilde{x}+\bm{\delta}), \tilde{y};\gamma\bm{\mathnormal{w}})$.
Because $f_{\gamma\bm{\mathnormal{w}}}$ is linear, we can easily determine the optimal adversarial direction in the feature space $\mathcal{Z}$ using $\nabla_{g}\tilde{\ell}(g(\tilde{x}+\bm{\delta}), \tilde{y};\gamma\bm{\mathnormal{w}})$.
Since the scale of the adversarial perturbation in the feature space is a problem of maximizing the convex function $\tilde{\ell}(g(\tilde{x}+\bm{\delta}), \tilde{y};\gamma\bm{\mathnormal{w}})$, as the scale of the perturbations increases, the situation is better from the adversarial point of view. However, these principles limit the scale range. By (\romannumeral 1), $\lambda_i>\eta=\abs{\mathop{\mathbb{E}}\left[\tilde{\mathnormal{z}}_i\right]}$, where $i\in\{2,\cdots,d+1\}$; by (\romannumeral 2), $\lambda_1<1=\abs{\mathop{\mathbb{E}}\left[\tilde{\mathnormal{z}}_1\right]}$.
Therefore, without loss of generality, the adversarial feature vector $\tilde{\bm{\mathnormal{z}}}^{\textup{adv}}$ can be approximated by $\tilde{\bm{\mathnormal{z}}}+\lambda\cdot\textup{sign}(\nabla_{\tilde{\bm{\mathnormal{z}}}}\tilde{\ell}(\tilde{\bm{\mathnormal{z}}}, \tilde{y}; \gamma\bm{\mathnormal{w}}))$ (we set $\eta<\lambda=\lambda_1=\cdots=\lambda_{d+1}<1$ for simplicity).

The loss function of the auxiliary task is formulated as
\begin{equation}
\begin{gathered}
\tilde{\ell}(\bm{\tilde{z}},\tilde{y};\gamma\bm{\mathnormal{w}})\\
=-\tilde{t}\ln{\sigma(\gamma\bm{\mathnormal{w}}^\top\tilde{\bm{\mathnormal{z}}}) - (1-\tilde{t})\ln{(1-\sigma(\gamma\bm{\mathnormal{w}}^\top\tilde{\bm{\mathnormal{z}}}))}},
\end{gathered}
\end{equation}
where $\tilde{t}=\frac{1}{2}(\tilde{y}+1)$. Therefore,
\begin{equation}\label{append2}
\begin{gathered}
\mathop{\mathbb{E}}\left[\tilde{\mathnormal{z}}^{\textup{adv}}_1\right]
=
\mathop{\mathbb{E}}\left[ \tilde{z}_1+\lambda\cdot\textup{sign}\left(\frac{\partial\tilde{\ell}}{\partial\tilde{z}_1}\right)\right]
\\=
y+\mathop{\mathbb{E}}\left[ \lambda\cdot\textup{sign}\left(\gamma\mathnormal{w}_1(\sigma(\gamma\bm{w}^\top\tilde{\bm{z}})-\tilde{t})\right) \right]=y,
\\
\mathop{\mathbb{E}}\left[\tilde{\mathnormal{z}}^{\textup{adv}}_i\right]
=
\mathop{\mathbb{E}}\left[ \tilde{z}_i+\lambda\cdot\textup{sign}\left(\frac{\partial\tilde{\ell}}{\partial\tilde{z}_i}\right)\right]
\\=
\eta y+\mathop{\mathbb{E}}\left[ \lambda\cdot\textup{sign}(\gamma\mathnormal{w}_i(\sigma(\gamma\bm{w}^\top\tilde{\bm{z}})-\tilde{t})) \right].
\end{gathered}
\end{equation}
We have
\begin{equation}\label{append3}
\begin{gathered}
\textup{sign}(\gamma\mathnormal{w}_i(\sigma(\gamma\bm{w}^\top\tilde{\bm{z}})-\tilde{t}))\\
=
\textup{sign}(\mathnormal{w}_i)\cdot\textup{sign}(\gamma\sigma(\gamma\bm{w}^\top\tilde{\bm{z}})-\gamma\tilde{t})=-y.
\end{gathered}
\end{equation}
Hence,
\begin{equation}
\mathop{\mathbb{E}}\left[\tilde{\mathnormal{z}}^{\textup{adv}}_i\right]
=
\eta y - \lambda y,
\end{equation}
where $i\in\{2,\cdots,d+1\}$ and $t=\frac{1}{2}(y+1)$.
\end{proof}
\setcounter{equation}{8}
\newtheorem{theorem-append}{Theorem}
\begin{theorem-append}
Let $\ell(;\bm{\mathnormal{w}})$ and $\tilde{\ell}(;\gamma\bm{\mathnormal{w}})$ be the loss functions of the primary and auxiliary tasks, respectively, and $t=\frac{1}{2}(y+1)$. When the auxiliary data are closely related to the primary data from the perspective of robust and non-robust features, i.e., $\abs{\gamma}=1$, the expectation of the gradient of $\tilde{\ell}$ with respect to $\tilde{\mathnormal{z}}^{\textup{adv}}_i:i\in\{2,\cdots,d+1\}$ is
\begin{equation}
\begin{gathered}
    \mathbb{E}\left[\frac{\partial\tilde{\ell}}{\partial\tilde{\mathnormal{z}}^{\textup{adv}}_i}\right]=
    \frac{\gamma}{d}\mathbb{E}\left[\sigma(\gamma\bm{\mathnormal{w}}^\top\tilde{\bm{\mathnormal{z}}}^{\textup{adv}})-\gamma\mathnormal{t}-\frac{1-\gamma}{2}\right]\\=
    \frac{1}{d}\mathbb{E}\left[\sigma(\bm{\mathnormal{w}}^\top\bm{\mathnormal{z}}^{\textup{adv}})-\mathnormal{t}\right]=
    \mathbb{E}\left[\frac{\partial\ell}{\partial{\mathnormal{z}}^{\textup{adv}}_i}\right].
\end{gathered}
\end{equation}
\end{theorem-append}
\setcounter{equation}{17}
\begin{proof}
The expectation of the gradient of $\tilde{\ell}$ with respect to $\tilde{\mathnormal{z}}^{\textup{adv}}_i:i\in\{2,\cdots,d+1\}$ is
\begin{equation}
    \begin{gathered}
    \mathbb{E}\left[\frac{\partial\tilde{\ell}}{\partial\tilde{\mathnormal{z}}^{\textup{adv}}_i}\right]
    =
    \mathbb{E}\left[\frac{\gamma}{d}(\sigma(\gamma\bm{w}^\top\tilde{\bm{z}}^{\textup{adv}})-\tilde{t})\right].
    \end{gathered}
\end{equation}
Based on Equation~\ref{append3}, we obtain
\begin{equation}
\begin{gathered}
\mathbb{E}\left[\frac{\gamma}{d}(\sigma(\gamma\bm{w}^\top\tilde{\bm{z}}^{\textup{adv}})-\tilde{t})\right]
\\=
\frac{\gamma}{d}\mathbb{E}\left[\sigma(\gamma\bm{\mathnormal{w}}^\top\tilde{\bm{\mathnormal{z}}}^{\textup{adv}})-\gamma\mathnormal{t}-\frac{1-\gamma}{2}\right]
\\=
\frac{1}{d}\mathbb{E}\left[\sigma(\bm{\mathnormal{w}}^\top\bm{\mathnormal{z}}^{\textup{adv}})-\mathnormal{t}\right].
\end{gathered}
\end{equation}
\end{proof}
\setcounter{equation}{10}
\begin{theorem-append}
Let $\tilde{\ell}\left(;\gamma\bm{\mathnormal{w}}\right)$ be the loss function of the auxiliary task. Then, if $\abs{\gamma}=1$, with high probability, the signs of $\tilde{\mathnormal{z}}^{\textup{adv}}_i:i\in\{2,\cdots,d+1\}$ and the auxiliary loss gradient with respect to $\tilde{\mathnormal{z}}^{\textup{adv}}_i$ are
\begin{equation}
    \textup{sign}\left(\tilde{\mathnormal{z}}^{\textup{adv}}_i\right)
    =-\gamma\mathnormal{q}
    =\textup{sign}\left(\frac{\partial\tilde{\ell}}{\partial\tilde{\mathnormal{z}}^{\textup{adv}}_i}\right).
\end{equation}
\end{theorem-append}
\setcounter{equation}{19}
\begin{proof}
The gradient of $\tilde{\ell}$ with respect to $\tilde{\mathnormal{z}}_i:i\in\{2,\cdots,d+1\}$ is
\begin{equation}
    \frac{\partial\tilde{\ell}}{\partial\tilde{z}_i}=\frac{\gamma}{d}(\sigma(\gamma\bm{w}^\top\tilde{\bm{z}})-r),\quad\textup{where}\:r=\frac{1}{2}(q+1).
\end{equation}
Therefore, the adversarial feature $\tilde{\mathnormal{z}}^{\textup{adv}}_i$ can be calculated as $\tilde{\mathnormal{z}}^{\textup{adv}}_i=\tilde{z}_i-\lambda\gamma q$. Because $\mathbb{E}\left[\tilde{z}_i\right]=\eta y$ and $\eta<\lambda$, the sign of $\tilde{\mathnormal{z}}^{\textup{adv}}_i$ is equal to $-\gamma q$ with high probability. In addition, the gradient of $\tilde{\ell}$ with respect to $\tilde{\mathnormal{z}}^{\textup{adv}}_i$ is given as
\begin{equation}
    \frac{\partial\tilde{\ell}}{\partial\tilde{\mathnormal{z}}^{\textup{adv}}_i}
    =
    \frac{\gamma}{d}(\sigma(\gamma\bm{w}^\top\tilde{\bm{z}}^\textup{adv})-r).
\end{equation}
Considering the adversarial vulnerability of our classification model, we can rewrite $\sigma(\gamma\bm{w}^\top \tilde{\bm{\mathnormal{z}}}^{\textup{adv}})$ as $\frac{1}{2} (1-\zeta\mathnormal{q})$, where $\zeta \in (0,1)$. Then,
\begin{equation}
    \frac{\partial\tilde{\ell}}{\partial\tilde{\mathnormal{z}}^{\textup{adv}}_i}
    =
    \frac{\gamma}{d}\left(\frac{1}{2}-\frac{\zeta q}{2}-\frac{q}{2}-\frac{1}{2}\right)
    =
    \frac{-\gamma q}{2d}(1+\zeta).
\end{equation}
Hence, the sign of $\frac{\partial\tilde{\ell}}{\partial\tilde{\mathnormal{z}}^{\textup{adv}}_i}$ is equal to $-\gamma q$ with high probability.
\end{proof}
\setcounter{equation}{11}
\begin{theorem-append}
Let $\tilde{\ell}(;\gamma\bm{\mathnormal{w}})$ be the loss function of the auxiliary task. Then, if $\abs{\gamma}=1$ and $\mathnormal{w}_1>0$, with high probability, the signs of $\tilde{\mathnormal{z}}^{\textup{adv}}_1$ and the auxiliary loss gradient with respect to $\tilde{\mathnormal{z}}^{\textup{adv}}_1$ are
\begin{equation}
    \textup{sign}(\tilde{\mathnormal{z}}^{\textup{adv}}_1)
    =\mathnormal{y},\quad
    \textup{sign}\left(\frac{\partial\tilde{\ell}}{\partial\tilde{\mathnormal{z}}^{\textup{adv}}_1}\right)
    =-\gamma\mathnormal{q}.
\end{equation}
\end{theorem-append}
\setcounter{equation}{22}
\begin{proof}
The gradient of $\tilde{\ell}$ with respect to $\tilde{\mathnormal{z}}_1$ is
\begin{equation}
    \frac{\partial\tilde{\ell}}{\partial\tilde{z}_1}=\gamma w_1(\sigma(\gamma\bm{w}^\top\tilde{\bm{z}})-r),\quad\textup{where}\:r=\frac{1}{2}(q+1).
\end{equation}
Assuming that the classification model is still vulnerable to adversarial examples, the adversarial feature $\tilde{\mathnormal{z}}^{\textup{adv}}_1$ is given as $\tilde{\mathnormal{z}}^{\textup{adv}}_1=\tilde{z}_1-\lambda\gamma q$. Because $\mathbb{E}\left[\tilde{z}_1\right]=y$ and $\lambda<1$, the sign of $\tilde{\mathnormal{z}}^{\textup{adv}}_i$ is equal to $y$ with high probability. In addition, the gradient of $\tilde{\ell}$ with respect to $\tilde{\mathnormal{z}}^{\textup{adv}}_1$ is
\begin{equation}
    \frac{\partial\tilde{\ell}}{\partial\tilde{\mathnormal{z}}^{\textup{adv}}_1}
    =
    \gamma w_1(\sigma(\gamma\bm{w}^\top\tilde{\bm{z}}^\textup{adv})-r).
\end{equation}
Considering the adversarial vulnerability of our classification model, $\sigma(\gamma\bm{w}^\top \tilde{\bm{\mathnormal{z}}}^{\textup{adv}})$ can be rewritten as $\frac{1}{2} (1-\zeta\mathnormal{q})$, where $\zeta \in (0,1)$. Then,
\begin{equation}
\begin{gathered}
    \frac{\partial\tilde{\ell}}{\partial\tilde{\mathnormal{z}}^{\textup{adv}}_1}
    =
    \gamma w_1\left(\frac{1}{2}-\frac{\zeta q}{2}-\frac{q}{2}-\frac{1}{2}\right)
    \\=
    \frac{-\gamma q w_1}{2}(1+\zeta).
    \end{gathered}
\end{equation}
Hence, the sign of $\frac{\partial\tilde{\ell}}{\partial\tilde{\mathnormal{z}}^{\textup{adv}}_1}$ is equal to $-\gamma q$ with high probability.
\end{proof}
If we use $\mathbb{E}\left[q\right]=0$ instead of sampled random labels $q$ for non-robust feature regularization, the gradient of $\tilde{\ell}$ with respect to $\tilde{\mathnormal{z}}_i:i\in\{2,\cdots,d+1\}$ is
\begin{equation}
    \frac{\partial\tilde{\ell}}{\partial\tilde{z}_i}=\frac{\gamma}{d}\left(\sigma(\gamma\bm{w}^\top\tilde{\bm{z}})-\frac{1}{2}\right).
\end{equation}
Based on the high standard accuracy of our classification model, with high probability, the gradient of $\tilde{\ell}$ with respect to $\tilde{\mathnormal{z}}_i:i\in\{2,\cdots,d+1\}$ can be rewritten as
\begin{equation}
\begin{gathered}
    \frac{\partial\tilde{\ell}}{\partial\tilde{z}_i}
    =
    \frac{\gamma}{d}\left(\sigma(\gamma\bm{w}^\top\tilde{\bm{z}})-\frac{1}{2}\right)
    \\=
    \frac{\gamma}{d}\left(\frac{1}{2}(1+\zeta\gamma y)-\frac{1}{2}\right)
    =
    \frac{\zeta y}{2d}.
\end{gathered}
\end{equation}
Therefore, the adversarial feature $\tilde{\mathnormal{z}}^{\textup{adv}}_i$ can be calculated as $\tilde{\mathnormal{z}}^{\textup{adv}}_i=\tilde{z}_i+\lambda y$. Because $\mathbb{E}\left[\tilde{z}_i\right]=\eta y$ and $\eta<\lambda$, the sign of $\tilde{\mathnormal{z}}^{\textup{adv}}_i$ is equal to $y$ with high probability. In addition, the gradient of $\tilde{\ell}$ with respect to $\tilde{\mathnormal{z}}^{\textup{adv}}_i$ is given as
\begin{equation}
    \frac{\partial\tilde{\ell}}{\partial\tilde{\mathnormal{z}}^{\textup{adv}}_i}
    =
    \frac{\gamma}{d}\left(\sigma(\gamma\bm{w}^\top\tilde{\bm{z}}^\textup{adv})-\frac{1}{2}\right).
\end{equation}
Because $\tilde{\mathnormal{z}}^{\textup{adv}}_i=\tilde{z}_i+\lambda y$, $\sigma(\gamma\bm{w}^\top \tilde{\bm{\mathnormal{z}}}^{\textup{adv}})$ can be approximated by $\frac{1}{2}(1+\gamma y)$. Then,
\begin{equation}
    \frac{\partial\tilde{\ell}}{\partial\tilde{\mathnormal{z}}^{\textup{adv}}_i}
    =
    \frac{y}{2d}.
\end{equation}
Hence, with high probability, the signs of $\tilde{\mathnormal{z}}^{\textup{adv}}_i$ and the auxiliary loss gradient with respect to $\tilde{\mathnormal{z}}^{\textup{adv}}_i$ are
\begin{equation}
    \textup{sign}(\tilde{\mathnormal{z}}^{\textup{adv}}_i)
    =\mathnormal{y}
    =
    \textup{sign}\left(\frac{\partial\tilde{\ell}}{\partial\tilde{\mathnormal{z}}^{\textup{adv}}_i}\right).
\end{equation}
\subsection{When $\abs{\gamma}<1$}
When $\abs{\gamma}<1$ (weak correlation), our theorems can be replaced as follows:
\begin{theorem-append}
Let $\ell(;\bm{\mathnormal{w}})$ and $\tilde{\ell}(;\gamma\bm{\mathnormal{w}})$ be the loss functions of the primary and auxiliary tasks, respectively, and $\hat{\gamma}=\textup{sign}(\gamma)$. Then, the sign of the expectation of the gradient of $\tilde{\ell}$ with respect to $\tilde{\mathnormal{z}}^{\textup{adv}}_i:i\in\{2,\cdots,d+1\}$ is
\begin{equation}\label{eq31}
\begin{gathered}
    \textup{sign}\left(\mathbb{E}\left[\frac{\partial\tilde{\ell}}{\partial\tilde{\mathnormal{z}}^{\textup{adv}}_i}\right]\right)\\=
    \textup{sign}\left(\mathbb{E}\left[\frac{\gamma\hat{\gamma}}{d}\sigma(\abs{\gamma}\bm{\mathnormal{w}}^\top\tilde{\bm{\mathnormal{z}}}^{\textup{adv}})-\mathnormal{t}\right]\right)=-y
    \\=\textup{sign}\left(\mathbb{E}\left[\frac{1}{d}\sigma(\bm{\mathnormal{w}}^\top\bm{\mathnormal{z}}^{\textup{adv}})-\mathnormal{t}\right]\right)
    \\=
    \textup{sign}\left(\mathbb{E}\left[\frac{\partial\ell}{\partial{\mathnormal{z}}^{\textup{adv}}_i}\right]\right).
\end{gathered}
\end{equation}
\end{theorem-append}
\begin{theorem-append}
Let $\tilde{\ell}(;\gamma\bm{\mathnormal{w}})$ be the loss function of the auxiliary task and $\hat{\gamma}=\textup{sign}(\gamma)$. Then, with high probability, the signs of $\tilde{\mathnormal{z}}^{\textup{adv}}_i:i\in\{2,\cdots,d+1\}$ and the auxiliary loss gradient with respect to $\tilde{\mathnormal{z}}^{\textup{adv}}_i$ are
\begin{equation}
    \textup{sign}(\tilde{\mathnormal{z}}^{\textup{adv}}_i)
    =-\hat{\gamma}\mathnormal{q}
    =\textup{sign}\left(\frac{\partial\tilde{\ell}}{\partial\tilde{\mathnormal{z}}^{\textup{adv}}_i}\right).
\end{equation}
\end{theorem-append}
\begin{theorem-append}
Let $\tilde{\ell}(;\gamma\bm{\mathnormal{w}})$ be the loss function of the auxiliary task and $\hat{\gamma}=\textup{sign}(\gamma)$. Then, if $\abs{\gamma}=1$ and $\mathnormal{w}_1>0$, with high probability, the signs of $\tilde{\mathnormal{z}}^{\textup{adv}}_1$ and the auxiliary loss gradient with respect to $\tilde{\mathnormal{z}}^{\textup{adv}}_1$ are
\begin{equation}
    \textup{sign}(\tilde{\mathnormal{z}}^{\textup{adv}}_1)
    =\mathnormal{y},\quad
    \textup{sign}\left(\frac{\partial\tilde{\ell}}{\partial\tilde{\mathnormal{z}}^{\textup{adv}}_1}\right)
    =-\hat{\gamma}\mathnormal{q}.
\end{equation}
\end{theorem-append}
The theorems in the cases of $\abs{\gamma}<1$ show that the scale of the correlation coefficient  does not change our main idea. Moreover, the training signals generated from the auxiliary task are weakened as $\abs{\gamma}$ approaches 0 (shown in Equation~\ref{eq31}). Note that we consider only a common robust and non-robust feature space between the primary and auxiliary data in our theoretical model. Therefore, negative transfer, induced by learning exclusive features of auxiliary tasks, cannot be described in our model.
\section{The effects of the use of more auxiliary datasets}\label{append:moreaux}
\input{Table/table6}
We investigate the effects of the use of more auxiliary datasets under the proposed method and provide the experimental results in Table~\ref{tab:table-6}. The results demonstrate that the use of more auxiliary datasets does not always lead to further improvements in adversarial robustness. The results on CIFAR-10 indicate that the use of both SVHN and CIFAR-100 results in a lower degree of robustness than that achieved by using CIFAR-100 alone. Likewise, leveraging a combination of ImageNet and Places365 leads to more vulnerable models than that utilizing only ImageNet. In other words, the relationship between the primary and auxiliary datasets is more important to the proposed method than the number of auxiliary datasets.

In fact, this result is a general phenomenon that can be easily observed even in non-adversarial setting. To show this, we conducted an additional test in which: (1) the CIFAR-10 training set was classified into datasets that contain 25000, 12500, and 12500 samples, namely cifar-A, cifar-B, and cifar-C, respectively. We added uniform noise to the cifar-C dataset to sparsify the information included in the cifar-C dataset; (2) a classifier (ResNet18) was then trained on cifar-A using cifar-B and cifar-C as extra datasets with a batch size of 128 and evaluated on the test set. The results in Table~\ref{tab:table-generalph} indicate that although cifar-B and cifar-C each result in performance improvement as an additional data set, the use of both cifar-B and cifar-C results in a test accuracy lower than that achieved by using cifar-B alone. We hypothesize that that this is because the density of information in the training dataset is more important than the total amount of information included in the training dataset in terms of the minibatch gradient descent. In other words, when DNNs are trained with a small batch size, the quality of each minibatch gradient is more important than the total amount of information in the dataset. To confirm this, we additionally run the abovementioned experiments with larger batch sizes; in fact, Table~\ref{tab:table-generalph} reveal that the use of both cifar-B and cifar-C results in a higher test accuracy than that achieved by using cifar-B alone in large batch settings.
\input{Table/table_generalph}
\section{Robust dataset analysis}\label{append:robustdataset}
\input{Table/table_robustdataset}
Ilyas et al.~\cite{not_bugs_are_features} generated a robust dataset containing only robust features (relevant to an adversarially trained model) to demonstrate their existence in images. In particular, they optimized: \[ \min_{\mathnormal{x}_r}\norm{g(\mathnormal{x}_r)-g(\mathnormal{x})}_2 \], where $\mathnormal{x}$ is the target image and $g$ is the feature embedding function.
They initialized $\mathnormal{x}_r$ as a different randomly chosen image from the training set. Thus, the robust dataset consists of optimized $\mathnormal{x}_r$--target label $y$ pairs.

To confirm robust feature learning through the application of the proposed method, we construct robust datasets from the AT and AT+BiaMAT models. We then normally train models from scratch on each robust dataset using the cross-entropy loss and list the results in Table~\ref{tab:table-robustdb}. As shown, the robust dataset developed using the model trained with the proposed method results in more accurate and robust models than those trained on the robust dataset of the baseline model. The proposed method thus enables neural networks to learn better robust features via inductive transfer between adversarial training on the primary and auxiliary datasets.

\section{Comparison with other related methods}\label{append:comp}
\paragraph{Semi-supervised learning.}
\input{Table/table4}
Carmon et al.~\cite{unlabeled_adv} and Stanforth et al.~\cite{are_labels_adv} proposed a semi-supervised learning technique by augmenting the training dataset with unlabeled in-distribution data.
The main difference between them and BiaMAT is the distribution of additional data leveraged. For instance, Carmon et al.~\cite{unlabeled_adv} collected in-distribution data of the CIFAR-10 dataset from 80 Million Tinyimages dataset~\cite{80mti} and used the unlabeled data with pseudo labels.
Carmon et al.~\cite{are_labels_adv} categorized CIFAR-10 into labeled and unlabeled data. Their theoretical analysis also assumed that the unlabeled data were in-distribution, and when out-of-distribution data were used instead, a large performance drop can be observed.
Therefore, while no assumptions are required for the classes of the primary and auxiliary datasets in our scenario, the semi-supervised methods are ineffective when the primary and auxiliary datasets do not share the same class distribution.
To demonstrate this, we assign pseudo labels to the auxiliary data using a classifier trained on each primary dataset and configure each training batch to contain the same amount of primary data and pseudo-labeled data as in \cite{unlabeled_adv}.
In particular, we sort the ImageNet data based on the confidence in the primary dataset classes and select the top ($N\times10$)k (or top ($N\times1$)k) samples for each class in CIFAR-10 (or CIFAR-100); this is denoted by ImageNet-($N\times100$)k.
In Table~\ref{tab:table-4}, the Carmon et al.~\cite{unlabeled_adv} method exhibits lower compatibility than the proposed method. In particular, the results obtained using CIFAR-100 and Places365 demonstrate that the semi-supervised method is vulnerable to negative transfer because of the considerable domain discrepancy between the primary and auxiliary datasets.
\paragraph{Pre-training.}
\input{Table/table5}
Hendrycks et al.~\cite{pretrain_hendrycks} demonstrated that ImageNet pre-training can significantly improve adversarial robustness on CIFAR-10. Although adversarial training on ImageNet is expensive, fine-tuning on the primary dataset does not require an extensive number of computations once the pre-trained model has been acquired.
However, once this has been done, it is difficult to obtain benefit from the application of cutting-edge methods in the fine-tuning phase because the hypothesis converges in the same basin in the loss landscape~\cite{what_is_being_transferred} when trained from pre-trained weights. For example, as shown in Table~\ref{tab:table-2}, TRADES generally achieves higher adversarial robustness than AT. However, fine-tuning a pre-trained ImageNet model~\cite{pretrain_hendrycks} through AT and TRADES, respectively, produces two models that exhibit similar levels of adversarial accuracy on CIFAR-10~(see Table~\ref{tab:table-5}).
By contrast, the proposed method can directly benefit from the application of state-of-the-art adversarial training methods~\cite{trades_defense,unlabeled_adv}. BiaMAT does not require complex operations and can also leverage a variety of datasets, whereas the pre-training method is effective only when a dataset that has a distribution similar to that of the primary dataset and a sufficiently large number of samples is used.
To demonstrate this difference empirically, we adversarially pre-train the CIFAR-100 and ImageNet models and then adversarially fine-tune them on CIFAR-10. The results in Table~\ref{tab:table-4} demonstrate that the pre-training method is ineffective when leveraging datasets that do not satisfy the conditions mentioned above. In other words, because the effect achieved by the pre-training method arises from the reuse of features pre-trained on a dataset that contains a large quantity of data with a distribution similar to that of the primary dataset, CIFAR-100 are not suitable for application of the CIFAR-10 task.
Conversely, BiaMAT avoids such negative transfer through the application of a confidence-based selection strategy. That is, these results emphasize the high compatibility of the proposed method with a variety of datasets.

\paragraph{Out-of-distribution data augmented training.}
Out-of-distribution data augmented training (OAT)~\cite{oat} was proposed as a means of supplementing the training data required for adversarial training. Under the assumption that non-robust features are shared among different datasets, the authors theoretically demonstrated that using out-of-distribution data with a uniform distribution label can reduce the contribution of non-robust features and empirically demonstrated that their method promotes the adversarial robustness of a model. 
OAT is similar to our proposed method in that it improves adversarial robustness by using additional data with a distribution that differs from that of the primary data. However, OAT does not derive useful information in terms of robust feature learning from auxiliary datasets. This is because OAT can only eliminate the contribution of features from the auxiliary dataset.
Therefore, BiaMAT outperforms OAT when the auxiliary dataset has a close relationship with the primary dataset in terms of robust features. By contrast, if the auxiliary dataset contains a large amount of useful information in terms of non-robust feature regularization rather than robust feature learning, the improvements resulting from the applications of OAT and BiaMAT can be similar.

BiaMAT has two advantages over OAT and RST:
\input{Table/tableEE}
(\romannumeral 1) OAT and RST assume that the given auxiliary dataset is out-distribution (OOD) and in-distribution (ID), respectively.
Hence, if a dataset contains both OOD and ID samples, they need an additional filtering process.
On contrary, BiaMAT is an end-to-end method that does not need any filtering;
(\romannumeral 2) If the assumptions on auxiliary datasets do not hold, OAT and RST will perform badly.
\Eg, OAT using ImageNet-100k (100k ImageNet samples closest to CIFAR-10) as an auxiliary dataset deteriorates the robustness on CIFAR-10. Tab.~\ref{tab:EE} indicates that in that case the BiaMAT model outperforms the OAT model by a large margin.

\paragraph{Generated data.}
Recently, Gowal et al. \cite{advddpm} leveraged generative models \cite{ddpm} to artificially increase the training dataset size. They showed that state-of-the-art robust accuracy can be achieved by using the increased training dataset. To be specific, they demonstrated that their proposed method yields the desired effect under the following conditions: (\romannumeral 1) The pre-trained non-robust classifier (pseudo-label generator) must be accurate on all realistic inputs. (\romannumeral 2) The generative model accurately approximate the true data distribution. 
From these conditions, we can infer the limitations of their method. That is, the effectiveness of their method is highly dependent on the quality of the generative and classification models that are solely trained on the original training dataset; in fact, Tab.~\ref{tab:table-4} demonstrates that the use of synthetic data leads to a significant robustness improvement on CIFAR-10 ($+3.69\%$), whereas a much smaller robustness improvement on CIFAR-100 ($+1.12\%$) than that induce by BiaMAT ($+3.05\%$).
In addition, Tab.~\ref{tab:table-4} shows that while \cite{advddpm} significantly improves robustness against AA, it has no effect on Clean. Based on these, we investigate whether the combination of \cite{advddpm} and BiaMAT, which considerably increases Clean, has a synergistic effect. Tab.~\ref{tab:B} indicates that BiaMAT can further improve \cite{advddpm}.

\input{Table/tableB}

\section{Implementation details}\label{append-implementation}
In all our experiments, we employed commonly used data augmentation techniques such as random cropping and flipping. On the CIFAR datasets, we used WRN28-10~\cite{wide_resnet} and WRN34-10 for AT and TRADES, respectively. On ImgNet100, we used WRN16-10.
\paragraph{Datasets.}
The CIFAR-10 dataset~\cite{cifar_dataset} contains 50K training and 10K test images over ten classes.
The CIFAR-100 dataset~\cite{cifar_dataset} includes 50K training and 10K test images over one hundred classes.
Each image in CIFAR-10 and CIFAR-100 consists of $32\times32$ pixels.
The ImageNet dataset~\cite{imagenet_dataset} has 1,281,167 training and 100,000 test images over 1,000 classes.
Chrabaszcz et al.~\cite{downsampled_imagenet} created downsampled versions of ImageNet. These datasets (ImageNet32x32 and ImageNet64x64)~\cite{downsampled_imagenet} contain the identical number of images and their classes as the original ImageNet dataset. The images therein are downsampled versions having pixel sizes of $32\times32$ and $64\times64$, respectively.
SVHN is obtained from a very large set of images from urban areas in various countries using Google Street View.
The CIFAR datasets are labeled subsets of the 80 million tiny images dataset~\cite{80mti}, and the 80 million tiny images dataset contains images downloaded from seven independent image search engines: Altavista, Ask, Flickr, Cydral, Google, Picsearch, and Webshots.
The Places365 images are queried from several online image search engines (Google Images, Bing Images, and Flickr) using a set of WordNet synonyms.
The ImageNet images are collected from online image search engines and organized by the semantic hierarchy of WordNet.

\paragraph{Training time.}
\input{Table/table_training_time}
The training times of the models are summarized in Tables~\ref{tab:table-training-time}. We used a single Tesla V100 GPU with CUDA10.2 and CuDNN7.6.5. Because of the increased training dataset size (and batch size) in the proposed method, the training time was almost twice that of the baseline method. Furthermore, a comparison of AT+BiaMAT(naive) and AT+BiaMAT revealed that the proposed confidence-based selection strategy requires negligible time.

\paragraph{Table~\ref{tab:table-1}.}
For the experiments in Table~\ref{tab:table-1}, we executed 100 training epochs on CIFAR-10. The initial learning rate was set to 0.1, and the learning rate decay was applied at 60\% and 90\% of the total training epochs with a decay factor of 0.1. Weight decay factor and $\ell_\infty\text-$bound were set to 2e-4 and $\frac{8}{255}$, respectively.

\paragraph{Table~2.} For the models associated with AT, we executed 100 training epochs (including 5 warm-up epochs) on CIFAR-10, CIFAR-100, and ImgNet100. The initial learning rate was set to 0.1, and the learning rate decay was applied at 60\% and 90\% of the total training epochs with a decay factor of 0.1. Weight decay factor and $\ell_\infty\text-$bound were set to 2e-4 and $\frac{8}{255}$, respectively.
Based on a recent study~\cite{bag_of_tricks_adv}, for the models associated with TRADES,  we executed 110 training epochs (including 5 warm-up epochs) on the CIFAR datasets and ImgNet100. The initial learning rate was set to 0.1, and the learning rate decay was applied at the 100th epoch and 105th epoch with a decay factor of 0.1. Weight decay factor and $\ell_\infty\text-$bound were set to 5e-4 and 0.031, respectively.

\input{Table/table9}
The hyperparameter $\alpha$ and $\pi$ for each model presented in Table~\ref{tab:table-1} is summarized in Table~\ref{tab:table-9}. From Table~\ref{tab:table-9}, it can be observed that when the proposed method is applied with AT, it produces good results around $\alpha=1.0$ and $\pi=0.5$ regardless of the primary dataset used. However, when the proposed method is applied with TRADES, the optimal set of hyperparameters are dependent on the characteristics of the primary task, such as the scale of training loss and its learning difficulty. For example, the primary task on CIFAR-10 achieves a lower training loss than that on CIFAR-100, and thus, a smaller $\alpha$ value is required when the primary dataset is CIFAR-10 than that required when the primary dataset is CIFAR-100. In addition, when the proposed method is applied to improve the sample complexity of a high-difficulty task, the confidence-based selection strategy becomes sensitive to the hyperparameter $\pi$, because the threshold used by the strategy is determined based on the confidences of the sampled primary data. Therefore, as a future research direction, we aim to develop an algorithm that can stably detect the data samples causing negative transfer.

When CIFAR-10 is the primary dataset, we use the same adversarial loss function for the primary and auxiliary tasks under BiaMAT. However, this setting can be problematic when the TRADES+BiaMAT model is trained on CIFAR-100.
TRADES uses the prediction of natural examples instead of labels to maximize the adversarial loss. In this respect, when an insufficient training time is applied to a challenging dataset, such as CIFAR-100 and ImageNet, low-quality training signals can arise owing to the inaccurate predictions. Therefore, in our experiment, the cross-entropy loss with labels is used for auxiliary tasks when the primary dataset is CIFAR-100. The application of the cross-entropy loss function allows the TRADES+BiaMAT models to achieve a high level of adversarial robustness on CIFAR-100, as shown in Table~\ref{tab:table-2}.

\paragraph{Pre-training.}
In the pre-training phase, the model was adversarially trained on the auxiliary dataset according to the implementation details described in Section~\ref{sec:expsetup}. The fine-tuning phase commenced from the best checkpoint of the pre-training phase. We adversarially fine-tuned the entire layers of the pre-trained model on the primary dataset.
The learning rate was set according to the global step over the pre-training and fine-tuning phase. For example, if the best checkpoint was acquired at the 65th epoch in the pre-training phase, the learning rate of the fine-tuning phase commenced at 0.01 and decreased to 0.001 after 25 epochs. When SVHN and CIFAR-100 were used as the auxiliary datasets, the abovementioned type of learning rate schedule rendered better robustness than that achieved by fine-tuning the model with a fixed learning rate~\cite{pretrain_hendrycks}.

\subsection{Ablation study on the hyperparameter $\pi$}\label{append:ablation}
\input{Table/table_ablation}
Here, we provide the results of ablation study on $\pi$ in Table~\ref{tab:table-ablation}. From the results of the AT+BiaMAT model, the effectiveness of BiaMAT is smooth near the optimal $\pi$ when it is applied with AT. In the results of TRADES+BiaMAT, however, it can be seen that the effectiveness of the proposed method is relatively sensitive to $\pi$ when it is applied with TRADES. We speculate that this is because of the relatively complex loss function of TRADES, which introduces another regularization hyperparameter $\beta$~\cite{trades_defense}. Therefore, in future work, we will develop advanced algorithms that adaptively control the threshold in BiaMAT for learning stability.

\input{Table/table_higher_than_thres}
\section{Additional analysis of the confidence-based selection strategy}\label{append:analysisconfidence}
Since robust features exhibit human-perceptible patterns, we conjecture that auxiliary data samples more related to the primary dataset classes can contribute more to robust feature learning. From this motivation, we design our algorithm to use the expectation of random labels for the less-related samples. In particular, we adopt an automatic confidence-based sample selection strategy, widely used in existing novelty detection literature \cite{hendrycks2018deep}. To understand how the proposed confidence-based selection strategy works in practice, we analyze the ratio of samples having higher confidences than the confidence threshold (\textit{i.e.}, $\omega$ in Algorithm \ref{algo}). If a sample contributes more to learn robust features, it tends to have a higher confidence score than less contributed samples.

We use the AT+BiaMAT model in Table \ref{tab:table-2}, trained on the CIFAR-10 dataset with the ImageNet auxiliary dataset. The model shows 88.75\% clean accuracy and 50.78\% robust accuracy on AA. Table \ref{tab:table_confidence} shows the average higher-than-threshold ratio (\textit{i.e}, the ratio of samples contribute to learn robust features) of ImageNet training images by the model. We show the average higher-than-threshold ratio for each CIFAR-10 superclasses.
We match classes of two datasets by using the ImageNet synset following CINIC-10 \cite{darlow2018cinic}\footnote{We follow the official synset mapping used by CINIC-10 \url{https://github.com/BayesWatch/cinic-10/blob/master/synsets-to-cifar-10-classes.txt}}.

\begin{figure}[htp]

\subfloat[The top-10 highest confident samples from ``aircraft carrier'' class]{%
  \includegraphics[clip,width=\columnwidth]{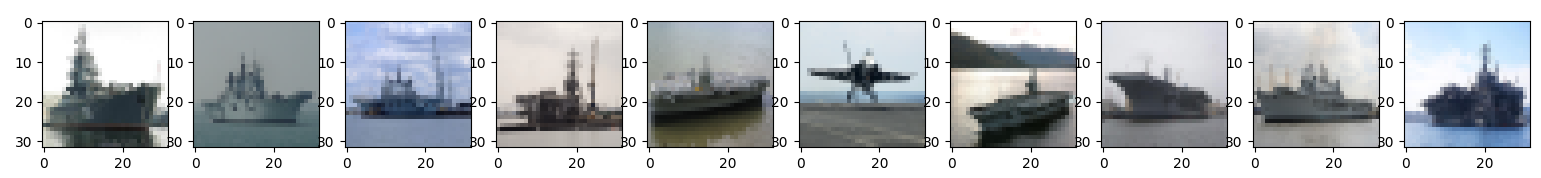}%
}

\subfloat[The top-10 lowest confident samples from ``aircraft carrier'' class]{%
  \includegraphics[clip,width=\columnwidth]{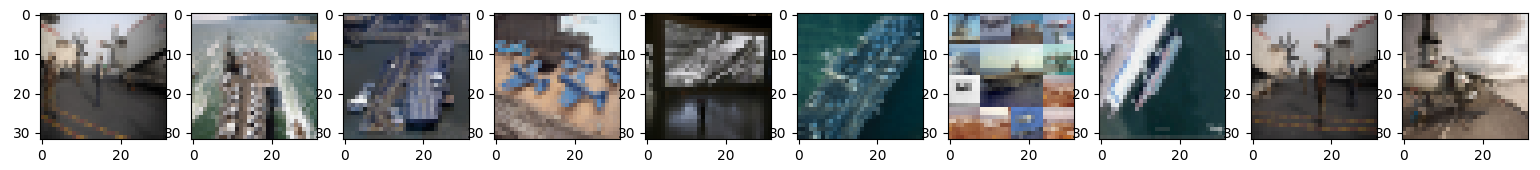}%
}

\caption{The top-10 highest and lowest confident ImageNet training samples (``aircraft carrier'' class) by the BiaMAT trained classifier on CIFAR-10}\label{fig:confidence_examples}

\end{figure}
In Table \ref{tab:table_confidence}, we observe that the related classes show higher selection ratio (larger than 50\%) than the dismatched classes (29\%) and the entire average (33.5\%). In other words, the auxiliary samples with CIFAR-10 superclasses contribute more to robust feature learning than less related samples (``Others'' in Table \ref{tab:table_confidence}). We also illustrate the samples from the class ``aircraft carrier'', showing 87.0\% higher-than-threshold ratio in Figure \ref{fig:confidence_examples}. In the figure, the highest confident samples plausibly match to the CIFAR-10 superclasses, such as ``Ship'' and ``Airplane''. On the other hand, the lowest confident samples, therefore their labels are shuffled during the training, seem to be less related to the CIFAR-10 superclasses and the original CIFAR-10 training images. The low confident samples can take a role of ``out-of-distributed'' dataset that can improve the confidence-based selection strategy as shown in \cite{hendrycks2018deep}.

Finally, we take a look into the ``Others'' classes as well. While the CIFAR-10 related classes show high higher-than-threshold ratios, we also witness that some classes not highly related to the CIFAR-10 superclasses, but weakly related to them also show high higher-than-threshold ratios. For example, (``grey whale'', 0.750), (``promontory'', 0.749), (``breakwater'', 0.734), (``dock'', 0.730), (``geyser'', 0.728), and (``sandbar'', 0.717) are not directly included in the CIFAR-10 superclasses, but share the similar environmental backgrounds (e.g., ``grey whale'' and ``ship'' are usually on the ocean background). The multi-domain learning strategy by BiaMAT let the model learn an auxiliary information by discriminating between such weakly related auxiliary classes and the CIFAR-10 superclasses. Our BiaMAT can learn better robust features by the additional tasks to discriminate weak auxiliary classes from the target classes.

To sum up, our confidence-based selection strategy let the model learn better robust features from plausible extra images, while less plausible images improve the performance of the confidence-based selection strategy. At the same time, the multi-domain learning strategy by BiaMAT makes the model learn discriminative features between the samples highly correlated with target classes and the sample weakly correlated with targets (e.g., ``grey whale''), thus BiaMAT shows a good robust feature learning capability. Therefore, BiaMAT can learn diverse and fine-grained features using extra images related to the target classes without suffering from the negative transfer, resulting in showing better robustness generalizability.

 From these observations, we conclude that by learning robust features from extra images but related to the primary dataset, a model can learn more diverse and fine-grained features, resulting in better robustness generalizability.

%% file: Table/table6.tex
\begin{table*}[]
\centering
\caption{Performance improvements~(accuracy \%) on CIFAR-10 following application of the proposed method using various datasets. The best result is indicated in bold.}
\begin{tabular*}{\textwidth}{c @{\extracolsep{\fill}} ccccc}
\toprule
Method                   & Auxiliary dataset        & Clean         & PGD\textsuperscript{$100$}          & CW\textsuperscript{$100$}    & AA         \\ \midrule[.1em]
                           \multirow{1}{*}{AT}   & -    & 87.37 & 50.87 & 50.93 & 48.53 \\\midrule[.05em]
                          \multirow{6}{*}{AT+BiaMAT} & SVHN & 87.34 & 51.90 & 51.40 & 48.61 \\
                          & CIFAR-100       & 87.22 & 55.93 & 52.09 & 50.08 \\
                          & SVHN, CIFAR-100       & 87.61 & 54.58 & 52.03 & 49.88 \\
                          & Places365     & 87.76 & 57.00 & 51.70 & 49.48 \\
                          & ImageNet      & \textbf{88.75} & \textbf{57.63} & \textbf{53.04} & \textbf{50.78}\\
                          & Places365,ImageNet     & 87.88 & 56.22 & 51.86 & 49.58
                           \\\midrule[.1em]
\end{tabular*}
\label{tab:table-6}
\end{table*}

%% file: Table/table_generalph.tex
\begin{table*}[]
\centering
\caption{Comparison (accuracy \%) of the effectiveness of data augmentation (cifar-B and cifar-C) on cifar-A.}
\begin{tabular*}{\textwidth}{c @{\extracolsep{\fill}} cc}
\toprule
Batch size                  & Dataset                   & Test error (mean$\pm$std over 5 runs)         \\ \midrule[.1em]
\multirow{4}{*}{128}  
                          & cifar-A  & 9.58$\pm$0.21 \\
                          & cifar-A + cifar-B  & \textbf{7.32}$\pm$0.14 \\
                          & cifar-A + cifar-C  & 9.15$\pm$0.26 \\
                          & cifar-A + cifar-B +cifar-C  & 7.45$\pm$0.21
                           \\\midrule[.05em]
\multirow{4}{*}{256}  
                          & cifar-A  & 10.48$\pm$0.21 \\
                          & cifar-A + cifar-B  & \textbf{8.06}$\pm$0.18 \\
                          & cifar-A + cifar-C  & 9.78$\pm$0.25 \\
                          & cifar-A + cifar-B +cifar-C  & 8.12$\pm$0.20
                           \\\midrule[.05em]
\multirow{4}{*}{384}  
                          & cifar-A  & 11.08$\pm$0.35 \\
                          & cifar-A + cifar-B  & 8.58$\pm$0.22 \\
                          & cifar-A + cifar-C  & 10.70$\pm$0.25 \\
                          & cifar-A + cifar-B +cifar-C  & \textbf{8.29}$\pm$0.21
                           \\\midrule[.05em]
\multirow{4}{*}{512}  
                          & cifar-A  & 11.49$\pm$0.20 \\
                          & cifar-A + cifar-B  & 9.22$\pm$0.12 \\
                          & cifar-A + cifar-C  & 11.21$\pm$0.27 \\
                          & cifar-A + cifar-B +cifar-C  & \textbf{8.94}$\pm$0.20
                           \\\midrule[.05em]
\multirow{4}{*}{1024}  
                          & cifar-A  & 13.22$\pm$0.25 \\
                          & cifar-A + cifar-B  & 10.55$\pm$0.21 \\
                          & cifar-A + cifar-C  & 12.85$\pm$0.33 \\
                          & cifar-A + cifar-B +cifar-C  & \textbf{10.23}$\pm$0.17
                           \\\midrule[.1em]
\end{tabular*}
\label{tab:table-generalph}
\end{table*}

%% file: Table/table_robustdataset.tex
\begin{table}[]
\centering
\caption{Accuracy (\%) comparison of the models~(WRN34-10) trained on each robust dataset generated from the AT and AT+BiaMAT models.}
\begin{tabular}{ccc}
\toprule
\multirow{2}{*}{Source model}     & \multirow{2}{*}{Clean}    & \multirow{2}{*}{\begin{tabular}[c]{@{}c@{}}FGSM\\ (mean$\pm$std over 5 runs)\end{tabular}}         \\ \\ \midrule[.1em]
\multirow{1}{*}{AT}  
                          & 87.49$\pm$0.20  &  30.79$\pm$1.16
                           \\\midrule[.05em]
\multirow{1}{*}{AT+BiaMAT}  
                          & \textbf{88.19}$\pm$0.16  &  \textbf{31.82}$\pm$1.06
                           \\\midrule[.1em]
\end{tabular}
\label{tab:table-robustdb}
\end{table}


%% file: Table/table4.tex
\begin{table*}[]
\centering
\caption{Comparison (accuracy \%) of the effectiveness of BiaMAT with the semi-supervised~\cite{unlabeled_adv} and pre-training~\cite{pretrain_hendrycks} methods on the CIFAR datasets.}
\begin{tabular*}{\textwidth}{c @{\extracolsep{\fill}} cccc}
\toprule
Primary dataset                  & Method                   & Auxiliary dataset        & Clean          & AA         \\ \midrule[.1em]
\multirow{11}{*}{CIFAR-10}  
                          & \multirow{2}{*}{Hendrycks et al.~\cite{pretrain_hendrycks}}   & CIFAR-100    & 80.21  & 42.36 \\
                          & & ImageNet     & 87.11  & 55.30
                          \\\cmidrule{2-5}
                          & \multirow{6}{*}{Carmon et al.~\cite{unlabeled_adv}} 
                          & CIFAR-100     & 82.61  & 50.81 \\
                          & & Places365     & 83.95  & 52.81 \\
                          & & ImageNet     & 85.42 &  53.79 \\
                          & & ImageNet-500k & 86.02 & 55.63 \\
                          & & ImageNet-250k & 86.51 & 56.27 \\
                          & & ImageNet-100k & 86.87 & 56.56 
                          \\\cmidrule{2-5}
                          & \multirow{1}{*}{Gowal et al.~\cite{advddpm}}   & Generated data \cite{ddpm}  & 85.07  & 57.62
                          \\\cmidrule{2-5}
                          & \multirow{3}{*}{\begin{tabular}[c]{@{}c@{}}TRADES+BiaMAT\\ (ours)\end{tabular}} 
                          & CIFAR-100     & 87.02  & 55.48 \\
                          & & Places365     & 87.18  & 55.24 \\
                          & & ImageNet     & 88.03 &  56.64
                           \\\midrule[.1em]
\multirow{8}{*}{CIFAR-100} 
                          & \multirow{1}{*}{Hendrycks et al.~\cite{pretrain_hendrycks}}  &ImageNet  & 59.23 & 28.79 \\\cmidrule{2-5}
                          & \multirow{5}{*}{Carmon et al.~\cite{unlabeled_adv}} & Places365     & 56.74 & 26.22  \\
                          & &ImageNet     & 63.45  & 27.71 \\
                          & &ImageNet-500k     & 64.90  & 28.64 \\
                          & &ImageNet-250k     & 66.18  & 29.49 \\
                          & &ImageNet-100k     & 65.40  & 30.61 
                          \\\cmidrule{2-5}
                          & \multirow{1}{*}{Gowal et al.~\cite{advddpm}}   & Generated data \cite{ddpm}  & 60.66  & 29.94
                          \\\cmidrule{2-5}
                          & \multirow{2}{*}{\begin{tabular}[c]{@{}c@{}}TRADES+BiaMAT\\ (ours)\end{tabular}} & Places365     & 64.58 & 29.24  \\
                          & &ImageNet     & 65.82  & 31.87 \\
                          
                          \midrule[.1em]
 
\end{tabular*}
\label{tab:table-4}
\end{table*}

%% file: Table/table5.tex
\begin{table}[]
\centering
\caption{Comparison~(accuracy \%) of the effectiveness of pre-training-based method using pre-trained ImageNet model on CIFAR-10 according to fine-tuning method}
\begin{tabular}{cccc}
\toprule
\multicolumn{1}{c}{Fine-tuning} & Clean & PGD20 & PGD100 \\ \midrule[.1em]
\multicolumn{1}{c}{AT} & 87.11 & 57.29 & 56.99 \\
\multicolumn{1}{c}{TRADES} & 83.97 & 57.17 & 57.07 \\
\bottomrule
\end{tabular}
\label{tab:table-5}
\end{table}

%% file: Table/tableEE.tex
\begin{table}[]
\centering
\caption{Results on CIFAR-10 when ImageNet-100k is auxiliary}
\label{tab:EE}
\begin{tabular}{ccccc}
\toprule
Method & Clean & AA\\\midrule
OAT \cite{oat}& 86.28 & 51.54\\
BiaMAT & \textbf{88.23} & \textbf{57.01}\\\bottomrule
\end{tabular}
\end{table}

%% file: Table/tableB.tex
\begin{table}[]
\centering
\caption{Performance improvements on CIFAR-10 (WRN16-8)} \label{tab:B}
{\footnotesize
\addtolength{\tabcolsep}{-2pt}
\begin{tabular}{cccccc}
\toprule
\multicolumn{3}{c|}{Clean} & \multicolumn{3}{c}{AA} \\
BiaMAT& \cite{advddpm}& \multicolumn{1}{c|}{BiaMAT+\cite{advddpm}}  & BiaMAT & \cite{advddpm} & BiaMAT+\cite{advddpm} \\\midrule

 84.51 & 82.68 & \multicolumn{1}{c|}{83.71} & 51.48 & 52.74 & 53.21\\\bottomrule

\end{tabular}
}
\end{table}

%% file: Table/table_training_time.tex
\begin{table}[t]
\centering
\caption{The training times of the models in our experiments.}
\begin{tabular}{c @{\extracolsep{\fill}} cc}
\toprule
Primary dataset                   & Method & Training time (h) \\ \midrule[.1em]
\multirow{5}{*}{CIFAR}  & AT  & 34 \\
& AT+BiaMAT (naive) & 56 \\
& AT+BiaMAT & 56.5\\
& TRADES & 52\\
& TRADES+BiaMAT & 103\\\midrule[.05em]
\multirow{2}{*}{ImgNet100}  & AT  & 119 \\
& AT+BiaMAT & 196
\\\midrule[.1em]
 
\end{tabular}
\label{tab:table-training-time}
\end{table}

%% file: Table/table9.tex
\begin{table*}[]
\centering
\caption{The hyperparameter $\alpha$ and $\pi$ for each model in Table~\ref{tab:table-2}}.
\begin{tabular*}{\textwidth}{c @{\extracolsep{\fill}} cccc}
\toprule
Primary dataset                  & Method                   & Auxiliary dataset        & $\alpha$ & $\pi$         \\ \midrule[.1em]
\multirow{7}{*}{CIFAR-10}  
                         
                          & \multirow{4}{*}{AT+BiaMAT} & SVHN & \multirow{4}{*}{1.0} & \multirow{4}{*}{0.55}\\
                          && CIFAR-100       &  &  \\
                          &  & Places365     &  & \\
                          &  & ImageNet & &\\\cmidrule{2-5}
                    
                          & \multirow{3}{*}{TRADES+BiaMAT} & CIFAR-100     & \multirow{3}{*}{0.5} & \multirow{3}{*}{0.5}\\
                          & & Places365     &  & \\
                          & & ImageNet &  & 
                           \\\midrule[.1em]
\multirow{4}{*}{CIFAR100} 
                          
                          & \multirow{2}{*}{AT+BiaMAT} &Places365 & \multirow{2}{*}{1.0} & \multirow{2}{*}{0.5}\\
                          && ImageNet  & & \\\cmidrule{2-5}
                          
                          & \multirow{2}{*}{TRADES+BiaMAT} & Places365     &\multirow{2}{*}{1.0} & \multirow{2}{*}{0.3}\\
                          & & ImageNet &  & \\\midrule[.1em]
\multirow{2}{*}{ImgNet100}     
                          
                          & \multirow{2}{*}{AT+BiaMAT} & Places365 & \multirow{2}{*}{1.0} & \multirow{2}{*}{0.5}\\
                          && ImgNet900           & & \\ \midrule[.1em]
 
\end{tabular*}
\label{tab:table-9}
\end{table*}

%% file: Table/table_ablation.tex
\begin{table}[]
\centering
\caption{The results of ablation study on $\pi$. Primary dataset: CIFAR-10; Auxiliary dataset: ImageNet.}
\begin{tabular}{ccc}
\toprule
Method   & $\pi$    & AA         \\ \midrule[.1em]
\multirow{6}{*}{AT+BiaMAT}  
                          & 0.45  &  49.85\\
                          & 0.50  & 50.35 \\
                          & 0.55 & 50.78 \\
                          & 0.60  & 50.32 \\
                          & 0.65  & 50.35 \\
                          & 0.70  & 50.69 
                           \\\midrule[.05em]
\multirow{6}{*}{TRADES+BiaMAT}  
                          & 0.45  &  56.42\\
                          & 0.50  & 56.64 \\
                          & 0.55 & 56.21 \\
                          & 0.60  & 54.70 \\
                          & 0.65  & 54.95 \\
                          & 0.70  & 54.04 
                           \\\midrule[.1em]
\end{tabular}
\label{tab:table-ablation}
\end{table}

%% file: Table/table_higher_than_thres.tex
\begin{table*}[t]
\centering
\caption{Average higher-than-threshold ratio of the ImageNet training images by the AT+BiaMAT-trained CIFAR-10 classifier. The fine-grained ImageNet classes are mapped to CIFAR-10 superclasses by the WordNet hierarchy. ``All'' denotes the entire training ImageNet images. ``Deer'' and ``Horse'' classes has zero error because there is only one ImageNet class matched to each of them (Table \ref{tab:table_confidence_classes}).}
\label{tab:table_confidence}
\begin{tabular}{@{}ccc@{}}
\toprule
CIFAR-10 Superclass  & Average higher-than-threshold ratio & Standard error \\ \midrule
Airplane    & 0.849              & 0.096          \\
Automobile  & 0.706              & 0.163          \\
Bird        & 0.554              & 0.143          \\
Cat         & 0.501              & 0.136          \\
Deer        & 0.720              & -          \\
Dog         & 0.592              & 0.103          \\
Frog        & 0.653              & 0.070          \\
Horse       & 0.819              & -          \\
Ship        & 0.677              & 0.215          \\
Truck       & 0.763              & 0.129          \\ \midrule
Others (dismatched) & 0.290              & 0.196          \\ \midrule
All & 0.335 & 0.219 \\ \bottomrule
\end{tabular}
\end{table*}

%% file: egpaper.bbl
\begin{thebibliography}{10}\itemsep=-1pt

\bibitem{adv_survey}
Naveed Akhtar and Ajmal Mian.
\newblock Threat of adversarial attacks on deep learning in computer vision: A
  survey.
\newblock {\em Ieee Access}, 6:14410--14430, 2018.

\bibitem{square_attack}
Maksym Andriushchenko, Francesco Croce, Nicolas Flammarion, and Matthias Hein.
\newblock Square attack: a query-efficient black-box adversarial attack via
  random search.
\newblock {\em arXiv preprint arXiv:1912.00049}, 2019.

\bibitem{cw_attack}
Nicholas Carlini and David Wagner.
\newblock Towards evaluating the robustness of neural networks.
\newblock In {\em 2017 IEEE Symposium on Security and Privacy (SP)}, pages
  39--57. IEEE, 2017.

\bibitem{unlabeled_adv}
Yair Carmon, Aditi Raghunathan, Ludwig Schmidt, John~C Duchi, and Percy~S
  Liang.
\newblock Unlabeled data improves adversarial robustness.
\newblock In {\em Advances in Neural Information Processing Systems}, pages
  11190--11201, 2019.

\bibitem{caruana1997multitask}
Rich Caruana.
\newblock Multitask learning.
\newblock {\em Machine learning}, 28(1):41--75, 1997.

\bibitem{Alvin}
Alvin Chan, Yi Tay, and Yew-Soon Ong.
\newblock What it thinks is important is important: Robustness transfers
  through input gradients.
\newblock In {\em Proceedings of the IEEE/CVF Conference on Computer Vision and
  Pattern Recognition (CVPR)}, June 2020.

\bibitem{simclr}
Ting Chen, Simon Kornblith, Mohammad Norouzi, and Geoffrey Hinton.
\newblock A simple framework for contrastive learning of visual
  representations.
\newblock In {\em International conference on machine learning}, pages
  1597--1607. PMLR, 2020.

\bibitem{downsampled_imagenet}
Patryk Chrabaszcz, Ilya Loshchilov, and Frank Hutter.
\newblock A downsampled variant of imagenet as an alternative to the cifar
  datasets.
\newblock {\em arXiv preprint arXiv:1707.08819}, 2017.

\bibitem{fab_attack}
Francesco Croce and Matthias Hein.
\newblock Minimally distorted adversarial examples with a fast adaptive
  boundary attack.
\newblock {\em arXiv preprint arXiv:1907.02044}, 2019.

\bibitem{auto_attack}
Francesco Croce and Matthias Hein.
\newblock Reliable evaluation of adversarial robustness with an ensemble of
  diverse parameter-free attacks.
\newblock {\em arXiv preprint arXiv:2003.01690}, 2020.

\bibitem{darlow2018cinic}
Luke~N Darlow, Elliot~J Crowley, Antreas Antoniou, and Amos~J Storkey.
\newblock Cinic-10 is not imagenet or cifar-10.
\newblock {\em arXiv preprint arXiv:1810.03505}, 2018.

\bibitem{imagenet_dataset}
J. Deng, W. Dong, R. Socher, L.-J. Li, K. Li, and L. Fei-Fei.
\newblock {ImageNet: A Large-Scale Hierarchical Image Database}.
\newblock In {\em CVPR09}, 2009.

\bibitem{multidomain2}
Mark Dredze, Alex Kulesza, and Koby Crammer.
\newblock Multi-domain learning by confidence-weighted parameter combination.
\newblock {\em Machine Learning}, 79(1-2):123--149, 2010.

\bibitem{visrobnonrob}
Logan Engstrom, Andrew Ilyas, Shibani Santurkar, Dimitris Tsipras, Brandon
  Tran, and Aleksander Madry.
\newblock Adversarial robustness as a prior for learned representations.
\newblock {\em arXiv preprint arXiv:1906.00945}, 2019.

\bibitem{fgsm_attack}
Ian~J Goodfellow, Jonathon Shlens, and Christian Szegedy.
\newblock Explaining and harnessing adversarial examples.
\newblock {\em arXiv preprint arXiv:1412.6572}, 2014.

\bibitem{advddpm}
Sven Gowal, Sylvestre-Alvise Rebuffi, Olivia Wiles, Florian Stimberg,
  Dan~Andrei Calian, and Timothy~A Mann.
\newblock Improving robustness using generated data.
\newblock {\em Advances in Neural Information Processing Systems}, 34, 2021.

\bibitem{pretrain_hendrycks}
Dan Hendrycks, Kimin Lee, and Mantas Mazeika.
\newblock Using pre-training can improve model robustness and uncertainty.
\newblock In {\em International Conference on Machine Learning}, pages
  2712--2721. PMLR, 2019.

\bibitem{outlier}
Dan Hendrycks, Mantas Mazeika, and Thomas Dietterich.
\newblock Deep anomaly detection with outlier exposure.
\newblock {\em arXiv preprint arXiv:1812.04606}, 2018.

\bibitem{hendrycks2018deep}
Dan Hendrycks, Mantas Mazeika, and Thomas Dietterich.
\newblock Deep anomaly detection with outlier exposure.
\newblock In {\em International Conference on Learning Representations}, 2019.

\bibitem{ddpm}
Jonathan Ho, Ajay Jain, and Pieter Abbeel.
\newblock Denoising diffusion probabilistic models.
\newblock {\em Advances in Neural Information Processing Systems},
  33:6840--6851, 2020.

\bibitem{not_bugs_are_features}
Andrew Ilyas, Shibani Santurkar, Dimitris Tsipras, Logan Engstrom, Brandon
  Tran, and Aleksander Madry.
\newblock Adversarial examples are not bugs, they are features.
\newblock {\em arXiv preprint arXiv:1905.02175}, 2019.

\bibitem{cifar_dataset}
Alex Krizhevsky, Geoffrey Hinton, et~al.
\newblock Learning multiple layers of features from tiny images.
\newblock Technical report, Citeseer, 2009.

\bibitem{avmixup}
Saehyung Lee, Hyungyu Lee, and Sungroh Yoon.
\newblock Adversarial vertex mixup: Toward better adversarially robust
  generalization.
\newblock In {\em Proceedings of the IEEE/CVF Conference on Computer Vision and
  Pattern Recognition (CVPR)}, June 2020.

\bibitem{oat}
Saehyung Lee, Changhwa Park, Hyungyu Lee, Jihun Yi, Jonghyun Lee, and Sungroh
  Yoon.
\newblock Removing undesirable feature contributions using out-of-distribution
  data.
\newblock In {\em International Conference on Learning Representations}, 2021.

\bibitem{pgd_attack}
Aleksander Madry, Aleksandar Makelov, Ludwig Schmidt, Dimitris Tsipras, and
  Adrian Vladu.
\newblock Towards deep learning models resistant to adversarial attacks.
\newblock {\em arXiv preprint arXiv:1706.06083}, 2017.

\bibitem{adaptive_multitask}
YUREN MAO, Weiwei Liu, and Xuemin Lin.
\newblock Adaptive adversarial multi-task representation learning.
\newblock In {\em Proceedings of Machine Learning and Systems 2020}, pages
  1832--1841. 2020.

\bibitem{multidomain1}
Hyeonseob Nam and Bohyung Han.
\newblock Learning multi-domain convolutional neural networks for visual
  tracking.
\newblock In {\em Proceedings of the IEEE conference on computer vision and
  pattern recognition}, pages 4293--4302, 2016.

\bibitem{transferability_adv}
Muhammad~Muzammal Naseer, Salman~H Khan, Muhammad~Haris Khan, Fahad~Shahbaz
  Khan, and Fatih Porikli.
\newblock Cross-domain transferability of adversarial perturbations.
\newblock In {\em Advances in Neural Information Processing Systems}, pages
  12885--12895, 2019.

\bibitem{svhn_dataset}
Yuval Netzer, Tao Wang, Adam Coates, Alessandro Bissacco, Bo Wu, and Andrew~Y
  Ng.
\newblock Reading digits in natural images with unsupervised feature learning.
\newblock 2011.

\bibitem{what_is_being_transferred}
Behnam Neyshabur, Hanie Sedghi, and Chiyuan Zhang.
\newblock What is being transferred in transfer learning?
\newblock In H. Larochelle, M. Ranzato, R. Hadsell, M.~F. Balcan, and H. Lin,
  editors, {\em Advances in Neural Information Processing Systems}, volume~33,
  pages 512--523. Curran Associates, Inc., 2020.

\bibitem{bag_of_tricks_adv}
Tianyu Pang, Xiao Yang, Yinpeng Dong, Hang Su, and Jun Zhu.
\newblock Bag of tricks for adversarial training.
\newblock In {\em International Conference on Learning Representations}, 2021.

\bibitem{rizve2021in}
Mamshad~Nayeem Rizve, Kevin Duarte, Yogesh~S Rawat, and Mubarak Shah.
\newblock In defense of pseudo-labeling: An uncertainty-aware pseudo-label
  selection framework for semi-supervised learning.
\newblock In {\em International Conference on Learning Representations}, 2021.

\bibitem{office}
Kate Saenko, Brian Kulis, Mario Fritz, and Trevor Darrell.
\newblock Adapting visual category models to new domains.
\newblock In {\em European conference on computer vision}, pages 213--226.
  Springer, 2010.

\bibitem{more_data_adversarial}
Ludwig Schmidt, Shibani Santurkar, Dimitris Tsipras, Kunal Talwar, and
  Aleksander Madry.
\newblock Adversarially robust generalization requires more data.
\newblock In {\em Advances in Neural Information Processing Systems}, pages
  5014--5026, 2018.

\bibitem{ssd}
Vikash Sehwag, Mung Chiang, and Prateek Mittal.
\newblock Ssd: A unified framework for self-supervised outlier detection.
\newblock {\em arXiv preprint arXiv:2103.12051}, 2021.

\bibitem{adv_transfer}
Ali Shafahi, Parsa Saadatpanah, Chen Zhu, Amin Ghiasi, Christoph Studer, David
  Jacobs, and Tom Goldstein.
\newblock Adversarially robust transfer learning.
\newblock In {\em International Conference on Learning Representations}, 2020.

\bibitem{fixmatch}
Kihyuk Sohn, David Berthelot, Nicholas Carlini, Zizhao Zhang, Han Zhang,
  Colin~A Raffel, Ekin~Dogus Cubuk, Alexey Kurakin, and Chun-Liang Li.
\newblock Fixmatch: Simplifying semi-supervised learning with consistency and
  confidence.
\newblock {\em Advances in Neural Information Processing Systems}, 33:596--608,
  2020.

\bibitem{are_labels_adv}
Robert Stanforth, Alhussein Fawzi, Pushmeet Kohli, et~al.
\newblock Are labels required for improving adversarial robustness?
\newblock {\em arXiv preprint arXiv:1905.13725}, 2019.

\bibitem{80mti}
Antonio Torralba, Rob Fergus, and William~T Freeman.
\newblock 80 million tiny images: A large data set for nonparametric object and
  scene recognition.
\newblock {\em IEEE transactions on pattern analysis and machine intelligence},
  30(11):1958--1970, 2008.

\bibitem{odds_with_accuracy}
Dimitris Tsipras, Shibani Santurkar, Logan Engstrom, Alexander Turner, and
  Aleksander Madry.
\newblock Robustness may be at odds with accuracy.
\newblock {\em arXiv preprint arXiv:1805.12152}, 2018.

\bibitem{overfitting_adv}
Eric Wong, Leslie Rice, and Zico Kolter.
\newblock Overfitting in adversarially robust deep learning.
\newblock In {\em Proceedings of Machine Learning and Systems 2020}, pages
  5304--5315. 2020.

\bibitem{wide_resnet}
Sergey Zagoruyko and Nikos Komodakis.
\newblock Wide residual networks.
\newblock {\em arXiv preprint arXiv:1605.07146}, 2016.

\bibitem{trades_defense}
Hongyang Zhang, Yaodong Yu, Jiantao Jiao, Eric Xing, Laurent~El Ghaoui, and
  Michael Jordan.
\newblock Theoretically principled trade-off between robustness and accuracy.
\newblock In Kamalika Chaudhuri and Ruslan Salakhutdinov, editors, {\em
  Proceedings of the 36th International Conference on Machine Learning},
  volume~97 of {\em Proceedings of Machine Learning Research}, pages
  7472--7482, Long Beach, California, USA, 09--15 Jun 2019. PMLR.
\newblock \url{https://github.com/yaodongyu/TRADES}.

\bibitem{places365}
Bolei Zhou, Agata Lapedriza, Aditya Khosla, Aude Oliva, and Antonio Torralba.
\newblock Places: A 10 million image database for scene recognition.
\newblock {\em IEEE Transactions on Pattern Analysis and Machine Intelligence},
  2017.

\end{thebibliography}
